\documentclass[11pt]{article}

\usepackage[final]{latex/acl}

\usepackage{times}
\usepackage{latexsym}
\usepackage{hyperref}
\usepackage[T1]{fontenc}

\usepackage[utf8]{inputenc}

\usepackage{microtype}

\usepackage{inconsolata}

\usepackage{graphicx}
\usepackage{subcaption}
\usepackage{amsmath}
\usepackage{amsfonts}
\usepackage{enumitem}
\usepackage{bm} 

\usepackage{booktabs} 
\usepackage{multirow} 
\usepackage{float}
\usepackage{makecell} 
\usepackage[normalem]{ulem}
\usepackage{tabularx}
\newcommand{\nama}[1]{\texttt{\textbf{#1}}}

\usepackage{tcolorbox}
\tcbuselibrary{breakable} 
\usepackage{tabularx}

\newtcolorbox{promptbox}[1][]{
  colback=gray!5,      
  colframe=gray!50,    
  coltitle=black,      
  boxrule=0.5pt,       
  arc=2mm,             
  fontupper=\ttfamily\small, 
  breakable,           
  #1
}

\usepackage{CJKutf8}
\title{Steering with Multilingual Lexicons for Sentiment Control in Low-Resource Languages}

\author{Lucinda Laurent$^{1}$ \quad Muhammad Dehan Al Kautsar$^{2}$ \\ \textbf{Alfan Farizki Wicaksono}$^{1}$  \quad \textbf{Fajri Koto}$^{2}$ \\ 
        $^{1}$Faculty of Computer Science, Universitas Indonesia \\
$^{2}$Mohamed bin Zayed University of Artificial Intelligence \\
	\texttt{\small lucinda.laurent@ui.ac.id, alfan@cs.ui.ac.id,
    \{muhammad.dehan,fajri.koto\}@mbzuai.ac.ae 
    } 
}
\begin{document}
\maketitle
\begin{abstract}

Activation steering offers a lightweight mechanism for controlling model behavior, but existing methods depend on sentence-level contrastive data that is unavailable for most of the world's languages. We show that multilingual sentiment lexicons are a viable substitute across three model families: aggregated multilingual lexicons outperform English-derived steering vectors for mid- and low-resource languages, while monolingual lexicons remain stronger for languages a model has specialized in during pre-training. The steering efficacy is mechanistically dictated by both representational density and pre-training priors. Large-magnitude English vectors overwhelm the compact representations of low-resource languages, causing ``linguistic collapse.'' This fragility is mitigated by the norm-dampening of multilingual aggregation. Architectural biases, including bilingual specialization penalty and latent numerical rigidity, also heavily constrain cross-lingual steering responsiveness. Ultimately, our work establishes lexicons as a practical steering signal and offers a comprehensive mechanistic account of cross-lingual steering dynamics.\footnote{Code and data are available at \url{https://github.com/lucindalaurent/SteeringWithMultilingualLexicon}.}

\end{abstract}

\section{Introduction}

Activation engineering has emerged as a lightweight method for controlling model outputs by manipulating internal activation vectors during inference, without modifying model weights or requiring curated training data \citep{turner:2024, rimsky:2024, arditi:2024, zou:2025}. By identifying linear directions associated with specific concepts (e.g., ``sentiment''), one can steer the model's behavior directly. In the context of sentiment-controlled generation, several studies have adopted activation engineering as an alternative to fine-tuning-based approaches~\citep{turner:2024,liu:2024,konen:2024}. Despite its promise, current sentiment steering methodologies are constrained by a reliance on sentence-level supervision and evaluation benchmarks optimized for English-centric corpora. This dependence creates a significant barrier for low-resource languages, where high-quality sentiment-labeled sentence corpora are scarce or non-existent, leaving most of the world's languages outside the reach of existing steering methods.

\begin{figure*}[t]
    \centering
    \includegraphics[width=0.95\textwidth]{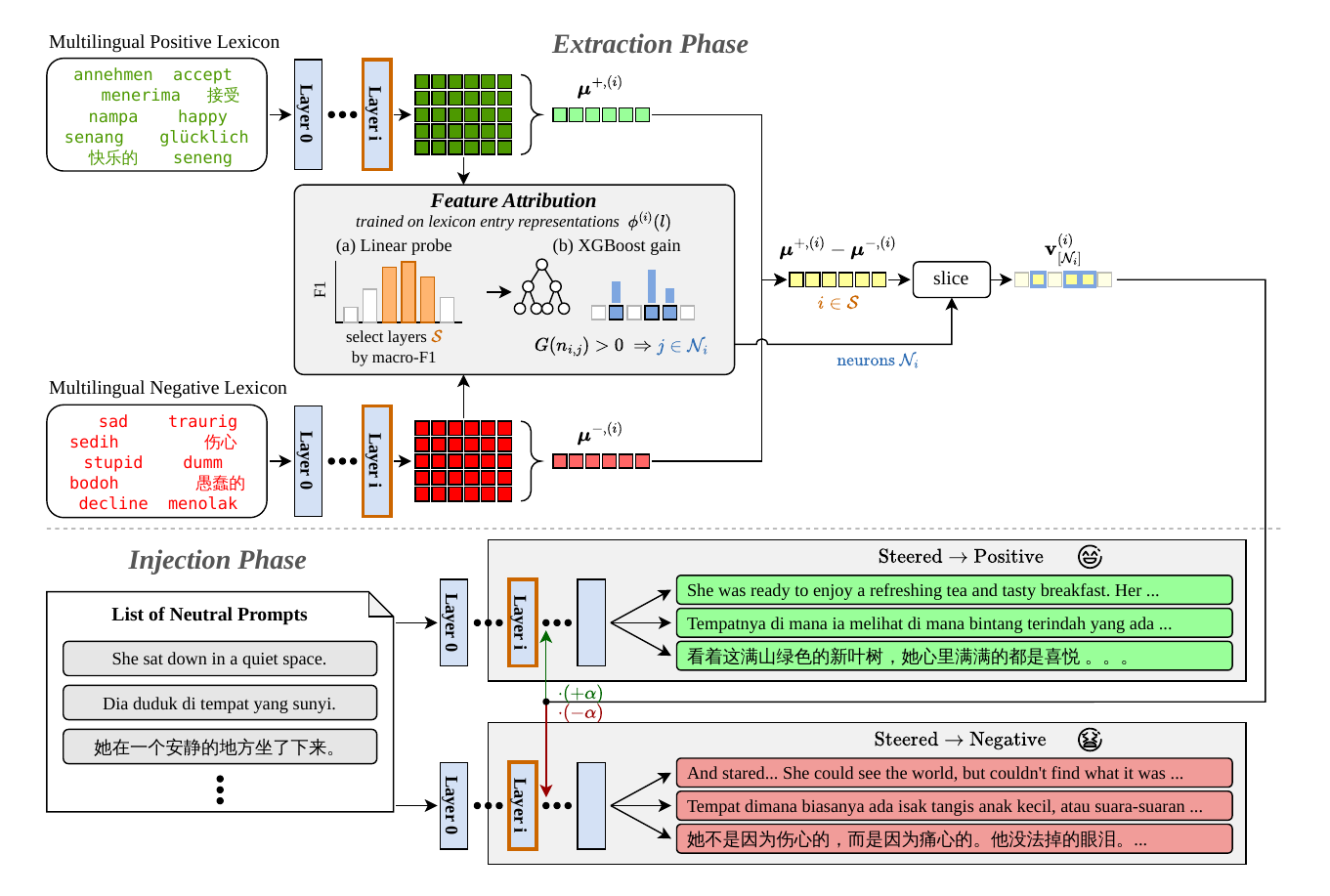}
  \caption{\textbf{Workflow of the Cross-Lingual Sentiment Control using Multilingual Lexicon.} In the Extraction Phase, feature attribution selects the steering layers $i \in \mathcal{S}$ (linear-probe macro-F1) and, within each, the neurons $\mathcal{N}_i$ (\nama{XGBoost} gain). At each selected layer, the steering vector is the difference $\mu^{+,(i)} - \mu^{-,(i)}$ between multilingual positive and negative lexicon centroids, sliced to $\mathcal{N}_i$. In the Injection Phase, this vector is applied to neutral prompts with coefficient $+\alpha$ or $-\alpha$, shifting the generated text toward positive or negative sentiment across multiple languages.}
    \label{fig:workflow}
\end{figure*}

The effectiveness of these interventions across diverse languages also remains largely unexplored. This gap is critical because LLMs exhibit varying degrees of cross-lingual isomorphism: the extent to which semantic concepts share a similar geometric representation across languages \citep{lample:2018}. In models with strong ``Anglocentric'' bias, a sentiment vector derived from English may be orthogonal to, or incongruent with, the sentiment subspace of low-resource languages like Javanese or Hausa, mirroring findings on the lack of isomorphism in low-resource cross-lingual embeddings \citep{sannigrahi:2022}. Relying solely on monolingual sentence-derived vectors thus constrains the scalability of activation steering in multilingual contexts.

In this work, we propose multilingual sentiment lexicons as a scalable, resource-efficient signal for cross-lingual activation steering. Our contributions are as follows:
\textbf{(1) A lexicon-based steering framework.} We formalize activation steering via multilingual sentiment lexicons (illustrated in Figure~\ref{fig:workflow}), eliminating the need for sentence-level annotations or paired contrastive prompts. 
\textbf{(2) Systematic cross-lingual evaluation.} We benchmark monolingual versus multilingual lexicon steering across $11$ typologically diverse languages spanning high-, mid-, and low-resource tiers and three model families. We show that aggregated multilingual lexicons outperform English-derived steering vectors for most mid- and low-resource languages, while monolingual lexicons remain superior in narrower cases where a model is heavily specialized in a particular language.
\textbf{(3) Mechanistic analysis of cross-lingual steering dynamics.} 
We demonstrate that steering success is dictated by a model's pre-training priors and latent scale disparities. We show that large-norm English vectors cause \textit{linguistic collapse} by overwhelming the compact latent subspaces of low-resource languages, whereas multilingual aggregation produces more balanced vectors that eliminate this mismatch and prevent representation collapse. Furthermore, we reveal how architectural biases determine a model's responsiveness to cross-lingual interventions.

\section{Background}

\paragraph{\nama{CTG} via Activation Steering.} 
Controllable Text Generation (\nama{CTG}) is the task of generating natural sentences that adhere to certain attributes \citep{prabhumoye:2020}. According to \citet{liang:2024}, \nama{CTG} methods can be classified based on the stage at which the intervention occurs: the training stage (e.g. retraining, fine-tuning, reinforcement learning) and the inference stage (e.g. prompt engineering, latent space manipulation, decoding-time intervention). Although training phase methods provide high-fidelity control, they are often constrained by high computational costs and substantial data requirements, making them impractical for low-resource language settings, where the extensive corpora required for effective training are unavailable. In this paper, we utilize the latent space manipulation (also known as activation engineering) method to exploit language model's underlying semantic geometry, potentially offering a more universal and robust framework for cross-lingual control. 

We specifically use the contrastive latent vector acquisition method to derive the steering direction (or steering vector). Several existing works have employed this approach for sentiment control, such as Activation Addition \citep{turner:2024}, In-Context Vectors \citep{liu:2024}, and Style Vectors \citep{konen:2024}. Activation Addition technique derived its steering vector by computing the difference of intermediate activations on a contrast pair of prompts, such as $h(\text{``Love''})-h(\text{``Hate''})$, where $h(w)$ represents the activation elicited by the word $w$. The $h(\text{``Love''})-h(\text{``Hate''})$ steering direction is then added during forward pass at inference time to shift the activations to generate positive sentiment completions. On the other hand, In-Context Vector (\nama{ICV}) is derived by applying Principal Component Analysis (\nama{PCA}) on the differences between hidden states from multiple example pairs $\{(x_i, y_i)\}_{i=1}^{n}$: 
\[
\nama{ICV} = \nama{PCA}\left(\{h(y_i) - h(x_i)\}_{i=1}^{n}\right) \, .
\]
In the demonstration for sentiment, the example pairs used are pairs of negative and positive text like (``Zero stars, I hate it.'', ``Five stars, I love it.''). 

Our method for deriving the sentiment steering direction is most similar to the Activation-based Style Vectors technique, in contrast to the other two approaches discussed. This technique extends the concept of Activation Addition by calculating a general style vector for a specific category $s \in S$. This is achieved by computing the difference between the mean activation of sentences in style category $s$ and the mean activation of all sentences in the remaining style categories, $S\setminus s$. In this study, instead of relying on labeled sentiment datasets such as \nama{Yelp} Review \citep{zhang:2015}, we utilize sentiment lexicons consisting of positive and negative words to address the inherent data scarcity in low-resource language settings.

\paragraph{Mechanistic Interpretability and Linear Representations.}
Rather than modifying model weights through fine-tuning, activation engineering exploits the pre-existing representations within the language model. It relies on the Linear Representation Hypothesis, assuming that semantic attributes are encoded as linear directions, thus allowing for direct steering within the representation space \citep{park:2024}. Previous research by \citet{tigges:2024} has shown that language models linearly represent sentiment, where methods like $k$-means, linear probing, and Distributed Alignment Search (\nama{DAS}), originally introduced by \citet{geiger:2024}, converge on a similar sentiment direction. \citet{dipalma:2025} further corroborates this by unveiling linearly probe-able sentiment and emotion representations specifically within the \nama{Llama} model family. We extend these findings by investigating the cross-lingual universality of sentiment directions; specifically, whether lexicon-derived sentiment directions align in a shared representation space across multiple languages. We primarily utilize the difference-in-means method to derive sentiment directions, complemented by linear probing to evaluate the consistency of sentiment representation across model layers. 

The findings by \citet{liuzhu:2024} suggest that in generative language models, lexical semantics are highly concentrated in the lower layers. By evaluating hidden states on the \nama{Word in Context (WiC)} benchmark, they demonstrate that lower layers manage contextual disambiguation (determining whether identical words convey the same meaning in distinct contexts), while higher layers shift computational focus toward next-token prediction. In contrast, our approach evaluates semantic polarity without surrounding context. We isolate the sentiment axis by extracting global representations directly from uncontextualized lexicon entries, which can range from single words to multi-word phrases. Because semantic encoding varies significantly across network depth \citep{liuzhu:2024}, mapping these isolated representations layer-by-layer allows us to empirically identify the optimal intervention point for cross-lingual sentiment steering using a multilingual lexicon.

\paragraph{Cross-Lingual Alignment and Universality.}
While the Linear Representation Hypothesis suggests stable sentiment directions within a single language, recent evaluations highlight that the generalization and reliability of steering vectors can degrade significantly under out-of-distribution shifts \citep{tan:2025}. Because cross-lingual transfer inherently represents a massive distributional shift, the capacity of these vectors to generalize zero-shot across languages remains a critical open question. Ideally, multilingual LLMs would exhibit cross-lingual isomorphism, where semantic spaces across languages share a similar geometric structure \citep{lample:2018}. This phenomenon was first empirically observed in massively multilingual encoders like \nama{mBERT} \citep{devlin:2019} and \nama{XLM-R} \citep{conneau:2020}, which demonstrated that pre-training on a shared vocabulary across $100+$ languages naturally induces a ``universal'' representation space where semantically similar tokens align across languages without explicit translation supervision \citep{libovicky:2020, pires:2019}. Under this assumption, a sentiment direction derived in English should theoretically transfer to other languages. 

Despite the assumption of a universal space, pre-training on unbalanced corpora frequently produces an ``Anglocentric'' internal structure \citep{ahuja:2023, lauscher:2020}. Within this structure, high-resource languages align closely with English subspaces, whereas mid-to-low resource languages suffer from weaker alignment or severe representation degeneration. This representational gap directly challenges cross-lingual steering; if the ``sentiment axis'' in a low-resource language is disjoint from the English axis, monolingual steering will fail. This geometric misalignment remains intrinsically tied to the model's underlying linguistic proficiency. Multilingual architectures rely on language-specific neurons to map inputs into the shared semantic space and decode them back into target vocabularies \citep{tang:2024}. Without robust neurons for low-resource languages, a severe projection bottleneck emerges, suggesting that the model's limited linguistic proficiency acts as a critical constraint on cross-lingual control, independent of the latent steering mechanism's geometric accuracy.

Operating under these constraints, we propose that aggregation of multilingual lexicons can serve as a geometric bridge. We hypothesize that if the underlying alignment is sufficiently preserved, a multilingual direction can reinforce the shared sentiment direction via coherent reinforcement. Conversely, in cases of extreme representation degeneration, such aggregation may fail to overcome the lack of semantic grounding.

\section{Methodology}
An overview of our cross-lingual sentiment steering framework is illustrated in Figure~\ref{fig:workflow}. 
\subsection{Sentiment Direction Extraction}
\label{subsec:direction_extraction}
Among the various methods for obtaining steering direction, difference-in-means is one of the most simple, yet effective approach \citep{marks:2024, belrose:2023}. To extract the steering direction corresponding to sentiment attribute, we employ the difference-in-means method using two opposing sets of lexicon entries (phrases), denoted as $\mathcal{L}^+$ and $\mathcal{L}^-$.

\paragraph{Definitions.}
Let $\mathcal{V}$ denote the model's vocabulary consisting of individual tokens. 
Recognizing that modern language models utilize sub-word tokenization (e.g., Byte-Pair Encoding), a single word or phrase in a lexicon may be decomposed into multiple tokens.
We therefore define a \textit{lexicon entry} $l$ as a sequence of tokens $l = (t_1, t_2, \dots, t_{|l|})$, where each $t_j \in \mathcal{V}$ and $|l| \ge 1$ represents the length of the entry. For example, the word ``abhorrence'' is tokenized into $(\text{``ab''}, \text{``hor''}, \text{``rence''})$ using \nama{Qwen-3}'s tokenizer. Accordingly, we define two contrastive sets of lexicon entries:
\begin{itemize}
    \item $\mathcal{L}^+ = \{l_1^+, l_2^+, \dots, l_N^+\}$ representing the positive lexicon set; and
    \item $\mathcal{L}^- = \{l_1^-, l_2^-, \dots, l_M^-\}$ representing the negative lexicon set.
\end{itemize}

\paragraph{Representation of Lexicon Entries.}
\label{subsec:lexicon_representation}



Let $h^{(i)}(t) \in \mathbb{R}^d$ denote the hidden state vector of a token $t$ at layer $i$. The vector representation of a lexicon entry $l$ consisting of tokens $(t_1, \dots, t_{|l|})$ at layer $i$, denoted as $\phi^{(i)}(l)$, is defined as the average of the hidden states of its constituent tokens at that specific layer:
\begin{equation*}
    \phi^{(i)}(l) = \frac{1}{|l|} \sum_{j = 1}^{|l|} h^{(i)}(t_j) \, .
\end{equation*}
By using these averaged hidden states as input, we ensure that the vector representation directly reflects the model's internal encoding of the sentiment-bearing terms in our lexicon. Furthermore, averaging across constituent tokens avoids over-reliance on the final token, which has been shown to be inconsistently informative for sentiment representation in decoder-only architectures, such as the \nama{Llama} family \citep{dipalma:2025}.

\paragraph{Layer-wise Centroids.}
We compute the centroid for the positive and negative groups by averaging the lexicon entry representations $\phi^{(i)}(l)$ strictly within the context of each layer $i$:
\begin{equation*}
\boldsymbol{\mu}^{+,(i)} = \frac{1}{|\mathcal{L}^+|} \sum_{l \in \mathcal{L}^+} \phi^{(i)}(l) \, ,\end{equation*} 
\begin{equation*}
    \boldsymbol{\mu}^{-,(i)} = \frac{1}{|\mathcal{L}^-|} \sum_{l \in \mathcal{L}^-} \phi^{(i)}(l) \, .
\end{equation*}
Here, $|\mathcal{L}^+|$ and $|\mathcal{L}^-|$ represents the number of lexicon entries in each set.

\paragraph{Layer-specific Steering Direction.}
Finally, the steering direction is computed independently for each layer $i$ over the full hidden dimension $d$. The steering direction is determined by the desired target attribute. To steer the model towards the positive attribute (enhancing $\mathcal{L}^+$ traits), the steering direction is defined as:
\begin{equation*}
    \mathbf{v}_{\nama{pos}}^{(i)} = \boldsymbol{\mu}^{+,(i)} - \boldsymbol{\mu}^{-, (i)} \, .
\end{equation*}
Conversely, to steer the model towards the negative attribute (or to suppress $\mathcal{L}^+$ traits), we reverse the direction: $\mathbf{v}_{\nama{neg}}^{(i)} = - \mathbf{v}_{\nama{pos}}^{(i)} $.

\subsection{Feature Attribution}
\label{subsec:feature_attribution}
In the context of activation steering, probing is frequently employed to identify the model components that contribute most significantly to the target attribute, thereby enabling more targeted interventions. For instance, \citet{li:2023} and \citet{wang:2025} utilize probes to select specific attention heads for steering, while \citet{konen:2024} employ probing to determine the optimal layers for intervention. Building upon these approaches, we propose a two-stage feature attribution mechanism: we first identify the layers with the highest sentiment density, and then isolate a sparse subset of neurons within those layers that contribute most significantly to the sentiment representation. 

Both stages are trained on the lexicon entry representations defined in Section~\ref{subsec:lexicon_representation}. For each layer, we construct a labeled dataset $\mathcal{D}^{(i)} = \{(\phi^{(i)}(l), y_l) : l \in \mathcal{L}^+ \cup \mathcal{L}^-\}$, where $y_l = 1$ if $l \in \mathcal{L}^+$ and $0$ otherwise.
$\mathcal{D}^{(i)}$ contains the same entries and labels for every layer $i$ and differs only in the layer at which the representations are taken. Since probing in isolation can capture non-causal correlations rather than true task relevance \citep{ravichander:2021, saphra:2024}, we conduct rigorous intervention experiments against random baselines, fully detailed in the Appendix~\ref{app:causal_verification}.
 
\paragraph{Layer-wise Probing.}
We hypothesize that sentiment information is non-uniformly distributed across the model's depth. To identify the most informative layers, we train a linear probe on $\mathcal{D}^{(i)}$ for each layer $i$ to predict the sentiment polarity $y \in \{0, 1\}$ using a logistic regression objective:
\begin{equation*}
    P(y \mid i, l) = \sigma\left[\mathbf{w}_i^T \phi^{(i)}(l) + b_i\right] \, ,
\end{equation*}
where $\mathbf{w}_i \in \mathbb{R}^d$ and $b_i \in \mathbb{R}$ are the learnable weights and bias of the probe, respectively. The layers are then ranked according to the probes' \nama{macro-F1} scores on the held-out portion of $\mathcal{D}^{(i)}$. Since the optimal subset of layers for intervention may vary across architectures and linguistic contexts, we consider three candidate layer configurations: all layers, the top $50\%$ of layers, and the top $5$ layers. For each combination of model, target language, and direction type (monolingual vs.\ multilingual), we use the configuration that yields the highest \textbf{Success Rate (SR)} after injection (Tables~\ref{tab:best_results_multi} and~\ref{tab:best_results_mono}). We denote the resulting set of selected layers by $\mathcal{S}$.

\paragraph{Sparse Neuron Selection via Non-linear Attribution.}
Steering all neurons within a selected layer may introduce unnecessary noise. Our preliminary experiments with \nama{Qwen} and \nama{Llama} indicate that this selection is an empirical necessity; unconstrained steering frequently resulting in incoherent, `gibberish' outputs. To mitigate this, we perform neuron-level attribution using Gradient Boosted Decision Trees (\nama{GBDT}) via \nama{XGBoost}. While sparsity-inducing linear probes (e.g., L1-regularized Logistic Regression) can reduce feature counts, they typically retain a broader set of neurons based on linear separability. In contrast, tree-based models perform a more aggressive, non-linear feature selection, isolating a highly compact subset of neurons highly discriminative for sentiment representation.

For each layer $i$, we train an \nama{XGBoost} classifier on $\mathcal{D}^{(i)}$, using the individual neurons of $\phi^{(i)}(l)$, denoted by $\{n_{i,1}, n_{i,2}, \dots, n_{i,d}\}$, as input features. We then calculate the \textit{gain} of each neuron $n_{i,j}$, denoted by $G(n_{i,j})$, which measures the average improvement in the training objective over all split points in the ensemble that use that neuron:
\begin{equation*}
    G(n_{i,j}) = \frac{1}{|\mathcal{P}_{i,j}|} \sum_{p \in \mathcal{P}_{i,j}} \Delta \ell_p \, ,
\end{equation*}
where $\mathcal{P}_{i,j}$ is the set of split points, across all trees, that split on neuron $n_{i,j}$; and $\Delta \ell_p$ is the improvement in the regularized log-loss objective achieved by split $p$, as computed by \nama{XGBoost}'s split criterion. We set $G(n_{i,j}) = 0$ for neurons that are never used in a split ($\mathcal{P}_{i,j} = \emptyset$). 
For each selected layer $i \in \mathcal{S}$, we define the active steering set $\mathcal{N}_i$ as the indices of the neurons with a non-zero gain ($G(n_{i,j}) > 0$).


\subsection{Sentiment Direction Injection}
\label{subsec:direction_injection}
We apply the derived steering direction during the inference phase by intervening on the model's internal activations. The intervention is performed at every forward pass of the model, specifically targeting the hidden states corresponding to the final token of the current sequence.


Let $\mathbf{x} = (x_1, x_2, \dots, x_t)$ be the input sequence at the current generation step $t$, where $x_t$ represents the last token (the most recently generated or processed token), and let $h^{(i)}(x_t)$ denote the original hidden state at the final token position for layer $i$.
For each selected layer $i \in \mathcal{S}$, we slice the full-dimensional steering direction $\mathbf{v}^{(i)}$, which can be either $\mathbf{v}_{\nama{pos}}^{(i)}$ or $\mathbf{v}_{\nama{neg}}^{(i)}$, to the active steering set $\mathcal{N}_i$, and add the scaled result to the corresponding neurons of the hidden state before it is passed to the subsequent layer $i+1$:
\begin{equation*}
    \tilde{h}^{(i)}_{[\mathcal{N}_i]}(x_t)
    \leftarrow
    h^{(i)}_{[\mathcal{N}_i]}(x_t)
    + \alpha \cdot \mathbf{v}^{(i)}_{[\mathcal{N}_i]} \, ,
\end{equation*}
where the subscript $[\mathcal{N}_i]$ denotes the sub-vector indexed by $\mathcal{N}_i$; $\tilde{h}^{(i)}(x_t)$ is the steered hidden state passed to the next layer; and $\alpha$ is the steering coefficient (hyperparameter) controlling the intervention strength. All neurons outside $\mathcal{N}_i$, as well as all layers $i \notin \mathcal{S}$, are left unchanged. 
By only intervening on the neurons in $\mathcal{N}_i$, we prevent the `gibberish' output phenomenon observed in unconstrained steering, thereby maintaining the structural integrity of the latent space while maximizing steering efficacy.


\section{Experiments}
\subsection{Experimental Setup}
\label{subsec:experimental_setup}

\paragraph{Models and Lexicon.} We evaluate three base LLMs: \nama{Llama-3.2-3B} \citep{llama:2025}, \nama{Gemma-3-4B-PT} \citep{gemma:2025}, and \nama{Qwen3-4B-Base} \citep{qwen:2025}. To derive steering directions, we utilize the \nama{NRC-VAD} lexicon \citep{nrc-vad:2018}, which provides human-rated valence scores $[0, 1]$. We apply strict thresholds ($> 0.8$ for $\mathcal{L}^+$ and $< 0.2$ for $\mathcal{L}^-$) to maximize semantic contrast and ensure robust steering vectors. We compare two configurations: \textbf{Monolingual} (English only) and \textbf{Multilingual}, which aggregates and deduplicates lexicons across 11 typologically diverse languages spanning three resource tiers: high-resource (English, German, Chinese, Japanese), mid-resource (Indonesian, Icelandic, Estonian, Basque), and low-resource (Hausa, Javanese, Sundanese). Detailed preprocessing and model configurations are provided in Appendix~\ref{app:experimental_details}.

\paragraph{Task, Dataset Construction, and Evaluations.} We evaluate cross-lingual steering on an open-ended generation task: given a neutral prompt, the model generates a continuation, and we measure whether steering shifts the sentiment of the generated text toward the target polarity while preserving the language of the prompt. To support this evaluation, we constructed a large-scale suite of $11{,}000$ neutral prompts across $11$ languages. We first generated $1{,}000$ unique neutral English prompts using a hierarchical structure ($10$ categories $\times$ $10$ topics $\times$ $10$ prompts per topic), then translated them into $10$ target languages using \nama{Gemini}. For low-resource languages, we employed style-guided prompts (e.g., specifying Javanese Ngoko Lugu and Sundanese Basa Loma) to strictly maintain a natural, conversational style. To ensure baseline reliability, we manually audited a $10\%$ random sample ($100$ prompts) of the unique English set (see Appendix~\ref{exp:dataset}). 


To assess steering, models generate continuations from these neutral points with a constraint of \textbf{at most $128$ new tokens}. This setup ensures sufficient context for sentiment analysis without forcing a fixed generation length. A separate tuning split of $50$ prompts was used for hyperparameter selection (see Appendix~\ref{app:prompts_and_topics} for topic lists and prompts). Continuations are evaluated using \nama{GPT-4.1}~\citep{gpt4} as a sentiment judge and \nama{GlotLID}~\citep{glotlid} for language identification. The use of large language models for automated text annotation has been shown to outperform crowd-workers across various classification tasks \citep{gilardi:2023}, with specific validation demonstrating high human-alignment in sentiment and emotion analysis \citep{calderon:2025, ruder:2025}. To confirm this reliability extends to our specific low-resource settings, we conducted native-speaker validation audits (detailed in Appendix~\ref{app:llm_judge_validation}).
Based on these automated evaluations, we report three metrics: (1) \textit{Sentiment Accuracy} (\nama{SA}) and (2) \textit{LID Rate} (\nama{LIDR}) as constituent components, and (3) \textit{Success Rate} (\nama{SR}), which serves as our primary metric for analysis. The \textit{Success Rate} represents a strict intersection, requiring a generation to satisfy \textbf{both} correct sentiment and correct language simultaneously. Steering intensity $\alpha$ was optimized via grid search on the tuning split (details in Appendix \ref{app:experimental_details}).

\section{Results and Analysis}
\subsection{Effectiveness of Cross-Lingual Steering}
\label{sec:steering_results}
\begin{figure*}[h] 
    \centering
    \includegraphics[width=0.95\textwidth]{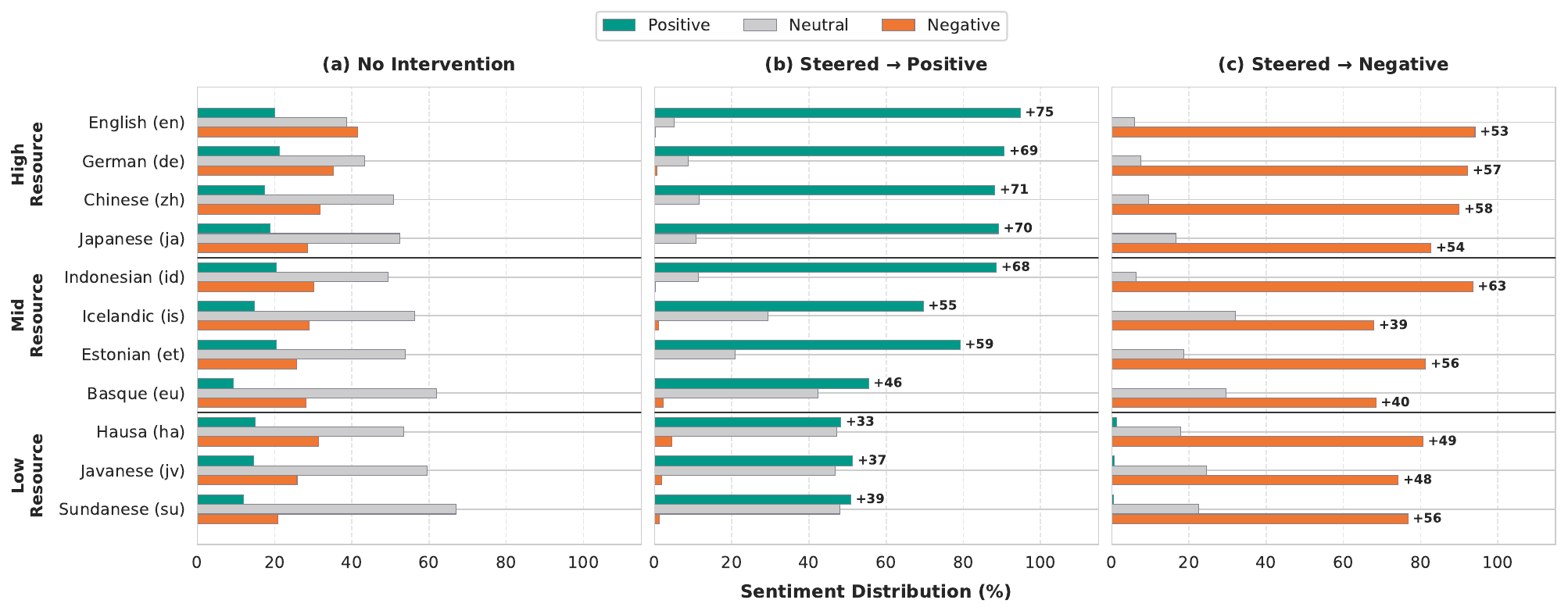}
    \caption{\textbf{Impact of Steering on Sentiment Distribution (Gemma-3).} The distribution of generated sentiments across 11 languages under three conditions: (a) No Intervention (as baseline), (b) Positive Steering, and (c) Negative Steering. The proposed method successfully shifts the predominantly neutral baseline toward the target sentiment. \textit{Corresponding distributions for} \nama{Qwen-3} \textit{and} \nama{Llama-3.2} \textit{are provided in Appendix \ref{app:additional_plots}.}}
    \label{fig:gemma_sentiment_dist}
\end{figure*}

\begin{table}[t]
\centering
\small
\setlength{\tabcolsep}{3.5pt}
\resizebox{\columnwidth}{!}{
\begin{tabular}{l c | cc | cc | cc}
\toprule
 & & \multicolumn{2}{c|}{\textbf{Gemma-3}} & \multicolumn{2}{c|}{\textbf{Qwen-3}} & \multicolumn{2}{c}{\textbf{Llama-3.2}} \\
\textbf{Lang} & \textbf{Res.} & \textbf{SR (\%)} & \textbf{Layers} & \textbf{SR (\%)} & \textbf{Layers} & \textbf{SR (\%)} & \textbf{Layers} \\
\midrule
English (en) & High & 94.3 & all & 45.1 & all & 65.8 & all \\
German (de) & High & 90.5 & all & 46.1 & top-50\% & 53.4 & all \\
Chinese (zh) & High & 86.2 & all & 48.0 & all & 33.1 & all \\
Japanese (ja) & High & 76.8 & all & 23.5 & all & 18.9 & all \\
\midrule
Indonesian (id) & Mid & 82.0 & all & 46.8 & all & 37.0 & all \\
Icelandic (is) & Mid & 66.8 & all & 10.9 & all & 23.1 & top-50\% \\
Estonian (et) & Mid & 77.8 & all & 23.5 & top-50\% & 25.4 & top-50\% \\
Basque (eu) & Mid & 60.7 & all & 12.2 & all & 20.4 & top-50\% \\
\midrule
Hausa (ha) & Low & 58.9 & top-50\% & \phantom{0}2.1 & top-5 & \phantom{0}9.6 & top-5 \\
Javanese (jv) & Low & 37.9 & top-50\% & \phantom{0}9.8 & top-50\% & \phantom{0}4.2 & top-50\% \\
Sundanese (su) & Low & 27.2 & all & \phantom{0}5.7 & top-50\% & \phantom{0}5.9 & top-50\% \\
\bottomrule
\end{tabular}
}
\caption{\textbf{Cross-Lingual Steering Performance with Multilingual Lexicons.} Success Rate (SR, \%) across three models and $11$ languages. Each cell reports SR at the best-performing layer configuration (all, top-$50$\%, or top-$5$).}
\label{tab:best_results_multi}
\end{table}
\begin{table}[t]
\centering
\small
\setlength{\tabcolsep}{3.5pt}
\resizebox{\columnwidth}{!}{
\begin{tabular}{l c | cc | cc | cc}
\toprule
 & & \multicolumn{2}{c|}{\textbf{Gemma-3}} & \multicolumn{2}{c|}{\textbf{Qwen-3}} & \multicolumn{2}{c}{\textbf{Llama-3.2}} \\
\textbf{Lang} & \textbf{Res.} & \textbf{SR (\%)} & \textbf{Layers} & \textbf{SR (\%)} & \textbf{Layers} & \textbf{SR (\%)} & \textbf{Layers} \\
\midrule
English (en) & High & 95.0 & all & 44.5 & all & 66.4 & all \\
German (de) & High & 90.7 & all & 45.4 & top-50\% & 53.8 & all \\
Chinese (zh) & High & 86.8 & all & 53.4 & all & 34.3 & all \\
Japanese (ja) & High & 80.2 & all & 23.9 & top-50\% & 20.8 & all \\
\midrule
Indonesian (id) & Mid & 82.3 & all & 42.9 & top-50\% & 38.6 & all \\
Icelandic (is) & Mid & 62.3 & top-50\% & 12.1 & top-50\% & 20.0 & all \\
Estonian (et) & Mid & 78.0 & top-50\% & 19.2 & top-50\% & 22.1 & top-50\% \\
Basque (eu) & Mid & 54.7 & top-50\% & \phantom{0}9.2 & top-50\% & 19.6 & all \\
\midrule
Hausa (ha) & Low & 54.6 & top-50\% & \phantom{0}1.8 & top-5 & \phantom{0}9.3 & top-50\% \\
Javanese (jv) & Low & 37.8 & top-50\% & 10.2 & top-50\% & \phantom{0}4.7 & all \\
Sundanese (su) & Low & 25.6 & top-50\% & \phantom{0}4.7 & top-50\% & \phantom{0}7.5 & all \\
\bottomrule
\end{tabular}
}
\caption{\textbf{Cross-Lingual Steering Performance with Monolingual English Lexicons.} Success Rate (SR, \%) across three models and $11$ languages. Each cell reports SR at the best-performing layer configuration (all, top-$50$\%, or top-$5$).}
\label{tab:best_results_mono}
\end{table}

We first evaluate the impact of our steering directions on the model's generation. Figure~\ref{fig:gemma_sentiment_dist} visualizes the sentiment distribution shifts for the \nama{Gemma-3} model across $11$ languages using the steering direction derived from the multilingual lexicon (see Figure~\ref{fig:gemma_mono_dist} for the corresponding distributions utilizing monolingual lexicon). With no intervention (Figure \ref{fig:gemma_sentiment_dist}a), the model generates continuations based on its intrinsic probability distribution. Since the prompts are semantically neutral (e.g., `He turned on the computer'), the resulting sentiments are naturally diverse or predominantly neutral. In contrast, applying the steering direction results in a dramatic distributional shift: Positive Steering (Figure \ref{fig:gemma_sentiment_dist}b) successfully aligns the majority of generations with positive sentiment, while Negative Steering (Figure \ref{fig:gemma_sentiment_dist}c) induces negative sentiment. These shifts indicate that lexicon-derived directions provide a viable basis for modulating sentiment across diverse languages. However, this responsiveness is not uniform across architectures. While the sentiment shift is nearly total in \nama{Gemma-3}, we observe varying degrees of sensitivity in \nama{Qwen-3} and \nama{Llama-3.2} (see Appendix~\ref{app:additional_plots}). In these models, although the intervention significantly increases the target sentiment, they maintain a higher proportion of neutral generations, particularly in lower-resource languages such as Hausa (\texttt{ha}) and Sundanese (\texttt{su}). This suggests that model-specific biases and language-specific representation density play a role in the effectiveness of lexicon-derived steering.

To quantify this effectiveness, Tables~\ref{tab:best_results_multi} and~\ref{tab:best_results_mono} summarize the performance of Multilingual and Monolingual (english) steering directions. First, both strategies significantly alter the model's behavior to align the output with the target sentiment compared to the uncontrolled baseline. Second, the Multilingual Lexicon aggregation emerges as a critical enabler for mid- and low-resource languages. This is most evident in \nama{Gemma-3}, where the Multilingual vector consistently outperforms the Monolingual counterpart in lower-resource settings. For instance, the Multilingual approach improves the success rate for Icelandic (\texttt{is}) from $62.3\%$ to $66.8\%$ ($+4.5\%$) and for Basque (\texttt{eu}) from $54.7\%$ to $60.7\%$ ($+6.0\%$). Even for Hausa (\texttt{ha}), we see an improvement of over $4$ percentage points ($54.6\%$ to $58.9\%$). 

Interestingly, in \nama{Qwen-3}, we observe a notable boost in Indonesian (\texttt{id}), rising from $42.9\%$ to $46.8\%$. This demonstrates that aggregating diverse lexicons creates a more robust, language-agnostic sentiment direction that helps bridge the representation gap in under-resourced languages. To ensure the robustness of these observed gains against stochastic decoding variance, we conducted rigorous statistical significance testing (Two-Sample Proportion Z-Tests) across all resource tiers. We find that the multilingual steering advantage is highly significant in Mid- and capable Low-resource regimes (see Appendix \ref{app:stats_significance}). While the Multilingual direction is generally superior for low-resource languages across all models, the Monolingual vector remains more effective for specific high-resource pairs, most notably \nama{Qwen-3} in Chinese (\texttt{zh}), where performance drops from $53.4\%$ to $48.0\%$ when using the Multilingual lexicon. In \nama{Llama-3.2}, the differences between Mono and Multi are often marginal in high-resource settings (e.g., $66.4\%$ vs $65.8\%$ for English). These variations suggest that the optimal steering strategy is heavily influenced by the model's pre-training dynamics, which we analyze further in Section \ref{subsec:dynamics}.

\subsection{Sensitivity and Robustness Analysis}
\label{subsec:sensitivity}
\begin{figure}[t!] 
    \centering
    \includegraphics[width=0.6\columnwidth]{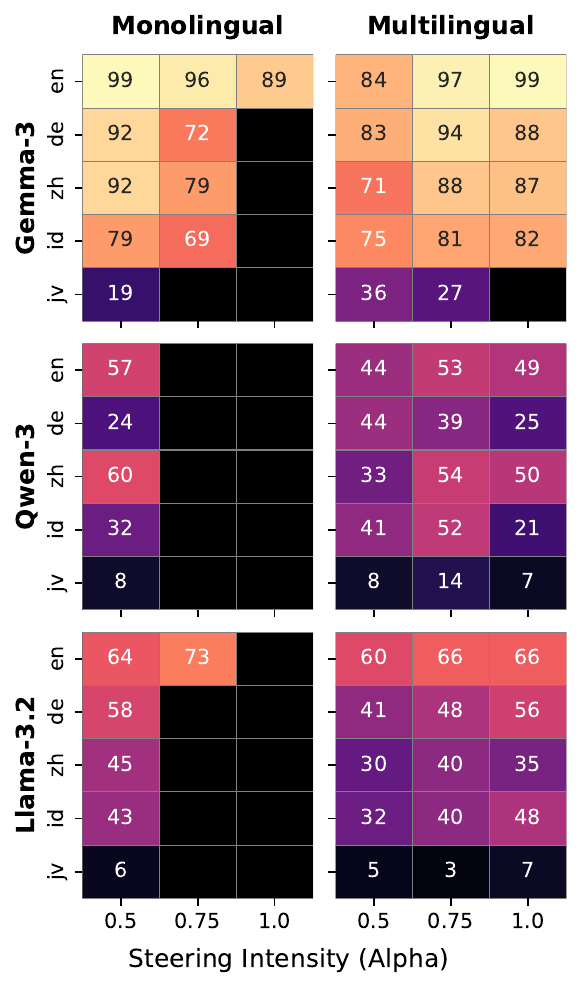}
   \caption{\textbf{Impact of Steering Intensity on Success Rate and Language Integrity.} 
    Heatmaps show performance under the all-layers configuration as the steering coefficient $\alpha$ increases from 0.5 to 1.0, for steering vectors derived from monolingual (English) and multilingual lexicons. 
    Values represent Success Rate (\%); black cells indicate linguistic collapse, where the generated text was incoherent and thus could not be evaluated for sentiment.}
    \label{fig:tuning_heatmap}
    \vspace{-0.5cm}
\end{figure}
To determine the optimal steering intensity and evaluate the stability of our vectors, we conducted a hyperparameter tuning sweep on the coefficient $\alpha \in \{0.50, 0.75, 1.00\}$. We focused this analysis on a representative subset of languages ranging from high- to low-resource: English (\texttt{en}), German (\texttt{de}), Chinese (\texttt{zh}), Indonesian (\texttt{id}), and Javanese (\texttt{jv}). Figure \ref{fig:tuning_heatmap} visualizes the results.

\begin{figure*}[h] 
    \centering
    \includegraphics[width=0.9\textwidth]{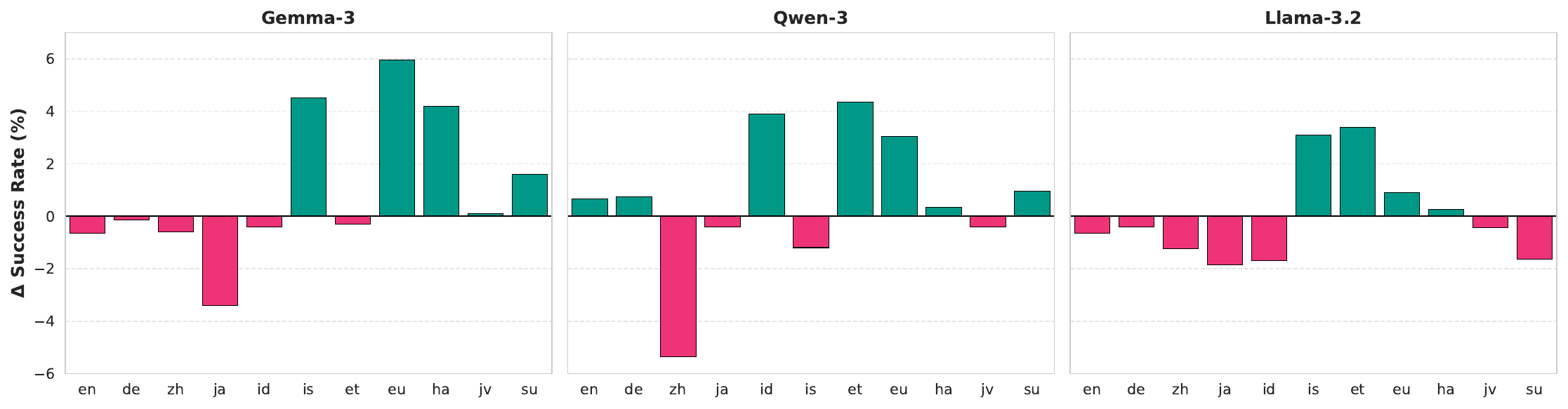}
    \caption{\textbf{Heterogeneous Impact of Multilingual Aggregation Across Model Families.} 
    This figure highlights the divergent impact of the Multilingual steering direction across three distinct model families. Positive values (teal) indicate that the Multilingual direction outperforms the Monolingual direction.}
    \label{fig:delta_plot}
    \vspace{-0.3cm}
\end{figure*}

\paragraph{Fragility of High-Norm Monolingual Steering Directions.}
The left column of Figure \ref{fig:tuning_heatmap} reveals that Monolingual steering is highly unstable at higher intensities. As $\alpha \to 1.0$, we observe widespread \textit{linguistic collapse} (black cells), particularly in \nama{Qwen-3} and \nama{Llama-3.2}, where the model output becomes incoherent. 
We hypothesize that this fragility stems from the high magnitude (norm) of the Monolingual steering direction. Since English sentiment representations are highly separable, the resulting steering direction encodes a ``sharp'' and distinct direction with a large norm. 
When this high-norm vector is injected into the latent space of other languages, especially low-resource ones like Javanese where representations are less distinct, it pushes the activations far out of the natural distribution, causing the generation to break down.

\paragraph{Robustness and Norm Dampening in Multilingual Steering Directions.}
In contrast, the Multilingual steering directions (right column) generally demonstrate superior stability. By aggregating lexicons from diverse languages, the resulting vector represents a ``consensus'' direction. We posit that this aggregation process often creates a dampening effect on the vector norm, particularly when the component languages have varying alignment. This strictly moderated norm allows for stronger steering intensities without shattering the model's linguistic integrity.

\paragraph{Power vs. Stability in \nama{Gemma-3}.}


\nama{Gemma-3} demonstrates a unique power-stability trade-off: while achieving superior steering shifts, it is prone to linguistic collapse at $\alpha=1.0$ (Figure \ref{fig:tuning_heatmap}). This fragility stems from a massive latent scale disparity: as measured across layers in Appendix~\ref{app:steering_norms}, \nama{Gemma-3} operates in a high-variance regime ($L_2 > 10,000$) compared to the compressed spaces of \nama{Llama-3.2} ($\sim 25$) and \nama{Qwen-3} ($\sim 1,150$). Within \nama{Gemma-3}, high-resource anchors (EN, ZH) define a ``norm ceiling,'' while low-resource languages like Javanese occupy a compact subspace (see Figure~\ref{fig:norms_comparison}). Consequently, the multilingual steering signal ($\sim 5,000$) is $2.5\times$ larger than the native Javanese norm ($\sim 2,000$). At maximum intensity, this disproportionate force shatters the linguistic integrity of lower-resource representations while remaining effective for high-norm languages. This suggests that universal steering requires intensity scaling relative to a language’s native representational density.

\subsection{Cross-Lingual Dynamics and Model Architecture}
\label{subsec:dynamics}
While Section \ref{subsec:sensitivity} analyzed the stability of the steering vectors, this section examines their relative effectiveness. Figure \ref{fig:delta_plot} illustrates the performance difference, defined as: 
\[\Delta = \nama{SR}_{\text{Multi}} - \nama{SR}_{\text{Mono}} \, ,
\]
where $\nama{SR}_{m}$ denotes the success rate for configuration $m \in \{\text{Multi}, \text{Mono}\}$ under its optimal layer configuration for that specific model and target language. Note that $\Delta$ is computed across $11$ languages. The patterns across the three models highlight how model architecture and pre-training priors dictate the optimal steering strategy.

\paragraph{Architectural Drivers of Steering Efficacy.} The patterns in Figure \ref{fig:delta_plot} reveal that a model’s pre-training priors dictate its responsiveness to multilingual aggregation. \nama{Gemma-3} exhibits the most consistent cross-lingual synergy, where multilingual vectors provide a coherent reinforcement signal for low-resource languages (e.g., \texttt{ha}, \texttt{su}) with minimal performance drops in high-resource languages: a benefit likely rooted in its optimized sub-word vocabulary. In contrast, \nama{Qwen-3} incurs a ``Specialization Penalty''; its bilingual-centric optimization for Chinese creates highly distinct representations that resist multilingual ``noise,'' leading to a $>5\%$ performance drop. For \nama{Llama-3.2}, multilingual aggregation serves as a critical bridge for mid-resource languages (e.g., \texttt{is}, \texttt{et}) where the English vector is semantically distant, yet it remains insufficient to overcome the baseline ``neutral anchor'' in the lowest-resource settings (\texttt{jv}, \texttt{su}). This resistance occurs despite high cross-lingual vector alignment (see Appendix~\ref{app:vector_alignment}), a paradox likely driven by the model's compressed latent scale and static layer-wise dynamics which create an internal numerical rigidity against external perturbations (see Appendix~\ref{app:steering_norms}).
\section{Conclusion}

We introduced multilingual sentiment lexicons as a scalable signal for cross-lingual activation steering, providing a practical alternative to sentence-level supervision where such data is unavailable. Across three model families and 11 languages, aggregated multilingual lexicons outperform English-derived steering vectors for mid- and low-resource languages, while monolingual lexicons remain preferable when a model is heavily specialized in a particular language. Our analysis further shows that successful intervention depends on internal model properties beyond surface-level similarity between sentiment directions: architectural specialization can decouple vector similarity from steering performance, and the scale mismatch between high-resource directions and compact low-resource representations can trigger linguistic collapse. These findings motivate a shift toward adaptive steering that calibrates intervention magnitude and lexicon configuration to each target language's representational scale.

\section*{Limitations}
\paragraph{Task and Model Scope.}
Our experiments were conducted exclusively on \textbf{open-ended sentence completion} tasks using pre-trained base models. We did not evaluate the efficacy of these steering vectors on instruction-following tasks (e.g., QA, summarization) or on instruction-tuned checkpoints (Chat models). Consequently, it remains unverified whether the observed steering dynamics persist when the model is constrained by strict instruction formatting or safety alignment filters. Furthermore, our analysis is limited to \textbf{efficient-scale models} in the 3B--4B parameter range. It is unclear if the trade-offs between stability and potency scale linearly to larger models (e.g., 70B+ parameters), where the latent space typically exhibits greater robustness and separability.
\paragraph{Generalization to Unseen Languages.}
The Multilingual directions in this study were aggregated using lexicons from a specific set of 11 languages. We evaluated the steering performance only on these in-domain languages and did not assess zero-shot transferability to unseen languages (languages whose lexicons were excluded from the aggregation). Evaluating on unseen languages would be an interesting extension and would further test the universality of the resulting directions. That said, the practical impact of this limitation is modest: sentiment lexicons for new languages can be obtained reliably through word-level translation, which is substantially more accurate and less error-prone than translating full sentences. Extending the multilingual lexicon set to cover additional languages is therefore straightforward in practice, even if the present results do not directly demonstrate zero-shot transfer.
\paragraph{Evaluation Metrics and Reliability.}
Our evaluation relied primarily on automated classifier-based metrics to measure sentiment shift and language consistency. 
First, we did not perform a deep quantitative analysis of textual fluency (e.g., perplexity) or topical relevance, meaning high sentiment scores do not guarantee that the generated text maintains semantic coherence or ``naturalness.'' 
Second, we acknowledge that the reliability of LLM-based evaluators (e.g., \nama{GPT-4}) degrades in low-resource settings. These models may harbor English-centric biases or lack the nuance to accurately adjudge sentiment in languages like Sundanese or Javanese. Consequently, the reported success rates in these languages reflect the evaluator's probability distribution but may not perfectly correspond to native-level perceptions of sentiment.

\section*{Ethics Statement}
\paragraph{Cultural Homogenization and Translation Artifacts.}
Our methodology relies on aggregating sentiment lexicons (\nama{NRC-VAD}) which are constructed via automated translation from English without recalibrating valence scores for the target cultures. Consequently, the resulting ``universal'' vectors inherently propagate an \textbf{Anglocentric definition of sentiment}. 
By projecting English sentiment values onto other languages, we risk enforcing a homogenized alignment where concepts deemed ``positive'' in Western norms override local cultural nuances (e.g., definitions of politeness or positivity in Javanese that differ from English). 
Users of this method should be mindful that the ``universality'' achieved here is mathematical rather than cultural; the method prioritizes cross-lingual robustness over cultural precision, potentially risking cultural erasure in low-resource languages by aligning them to a foreign sentiment standard.
\paragraph{Dual-Use and Safety.}
While this study aims to improve cross-lingual control, activation steering is a dual-use technology. The same mechanisms used to enforce positive sentiment can be inverted to amplify toxicity or hate speech. Furthermore, as noted in our limitations, our current evaluation does not guarantee semantic coherence. Deploying such steering methods in production without rigorous fluency and safety checks could lead to the generation of superficially ``positive'' but nonsensical or hallucinated content, potentially degrading user trust.

\paragraph{Artifact and License Compliance.}
All artifacts used in this study are restricted to their intended research purposes. The code libraries are utilized under standard open-source licenses (e.g., MIT, Apache 2.0), and all datasets and pre-trained models are accessed in strict accordance with their original creators' terms of use.

\bibliography{latex/custom}

\clearpage  
\raggedbottom
\appendix



\section{Experimental Details}
\label{app:experimental_details}

\subsection{Model and Implementation}
All models (\nama{Llama-3.2-3B}, \nama{Gemma-3-4B-PT}, and \nama{Qwen3-4B-Base}) were implemented using the \texttt{transformers} library \citep{wolf:2020} in \texttt{bfloat16} precision. We specifically utilize the \textbf{base (pre-trained)} versions to investigate sentiment representations in the foundational latent space, thereby avoiding biases from instruction-tuning or RLHF. Because all evaluated models are in the $\sim$$3$B to $4$B parameter range, they require approximately $16$ GB of VRAM for inference and activation steering. We distributed the experiments across several computing environments based on resource availability: publicly available cloud notebooks (dual NVIDIA T4, $2 \times 16$\,GB), commercial cloud GPU instances (NVIDIA RTX A4500, 20\,GB), and a shared academic compute cluster (NVIDIA DGX-A100).

\subsection{Data and Lexicon Preprocessing}
\paragraph{Lexicon Settings.} As discussed in Section \ref{subsec:direction_extraction}, we specifically utilize the multilingual \nama{NRC-VAD} lexicon \citep{nrc-vad:2018} to obtain the sentiment direction. The \nama{NRC-VAD} lexicon provides human ratings for valence, arousal, and dominance, with scores ranging from $0$ to $1$. In our experiments, we investigate the efficacy of cross-lingual steering by comparing two distinct lexicon configurations: \textit{monolingual} and \textit{multilingual}.

For the \textit{monolingual} configuration, we rely exclusively on English lexicon entries. The preprocessing involves removing invalid entries (\nama{NaN} values) and filtering the remaining entries based on their valence scores. Entries with a valence score $> 0.8$ are assigned to the positive set $\mathcal{L}^+$, while entries with a valence score $< 0.2$ are assigned to the negative set $\mathcal{L}^-$. We deliberately employ these strict thresholds to maximize the semantic contrast between the opposing sets, ensuring that the resulting steering direction is robust and not diluted by neutral or weakly polarized terms.

For the \textit{multilingual} configuration, we aggregate lexicon entries from 11 languages, selected to cover a spectrum from high-resource to low-resource languages: English (\texttt{en}), German (\texttt{de}), Chinese (\texttt{zh}), Japanese (\texttt{ja}), Indonesian (\texttt{id}), Icelandic (\texttt{is}), Estonian (\texttt{et}), Basque (\texttt{eu}), Hausa (\texttt{ha}), Javanese (\texttt{jv}), and Sundanese (\texttt{su}). The preprocessing for the aggregated set involves a rigorous deduplication process. First, we identify and remove ambiguous duplicates: entries appearing in multiple languages that exhibit conflicting valence scores (i.e., one instance having a score $< 0.2$ and another $> 0.8$). Second, for duplicates with consistent sentiment polarity (both $< 0.2$ or both $> 0.8$), we retain only the first occurrence. From this refined set of unique multilingual entries, we construct $\mathcal{L}^+$ and $\mathcal{L}^-$ using the same valence thresholds ($> 0.8$ and $< 0.2$) as the \textit{monolingual} configuration.

\paragraph{Dataset Construction and Validation}
\label{exp:dataset}
We constructed a multilingual evaluation suite consisting of 1000 neutral prompts across 11 languages. We utilized \nama{Gemini} to generate 10 categories and 100 sub-topics, which served as the foundation for creating 1,000 neutral evaluation sentences. The full list of categories and prompts is provided in Appendix \ref{app:prompts_and_topics}. These were subsequently translated into 10 target languages, with specialized prompts for low-resource languages (Hausa, Javanese, Sundanese) to ensure linguistic naturalness. To ensure the neutrality of the baseline, we manually audited a random sample of 100 prompts. While the majority were neutral, we observed that approximately 3\% contained implicit sentiment (e.g.,``The departure board showed the flight was ‘delayed’'' or ``A client reported an issue with the software''). These instances highlight that while \nama{Gemini} generates objectively factual statements, certain real-world contexts carry inherent negative or positive connotations.



\setlength{\tabcolsep}{0.2em}
\begin{table}[ht]
\centering
\small
\begin{tabular}{@{}lcccc@{}}
\toprule
\textbf{Lang.} & \textbf{ISO} & \textbf{Llama-3.2} & \textbf{Qwen3-4B} & \textbf{Gemma3-4B} \\ \midrule
English & en & $\mathbf{8.47}$ & $8.53$ & $8.54$ \\
German & de & $12.39$ & $12.42$ & $\mathbf{10.80}$ \\
Chinese & zh & $9.75$ & $\mathbf{7.62}$ & $8.03$ \\ 
Japanese & ja & $13.41$ & $12.43$ & $\mathbf{9.84}$ \\
\midrule
Indonesian & id & $12.51$ & $12.61$ & $\mathbf{9.41}$ \\
Icelandic & is & $17.42$ & $17.62$ & $\mathbf{15.14}$ \\
Estonian & et & $13.33$& $13.44$ & $\mathbf{11.78}$ \\
Basque & eu & $15.44$ & $15.61$ & $\mathbf{13.53}$ \\
\midrule
Hausa & ha & $15.51$ & $15.22$ & $\mathbf{13.68}$ \\
Javanese & jv & $14.35$ & $14.49$ & $\mathbf{12.64}$ \\
Sundanese & su & $15.02$ & $15.16$ & $\mathbf{12.54}$ \\ \bottomrule
\end{tabular}
\caption{Average tokens per neutral prompt across target languages. Lower values indicate higher tokenization efficiency and a more optimized vocabulary for that language.}
\label{tab:token_stats}
\end{table}

\subsection{Steering Methodology Details}
\paragraph{Linear Probing.} For layer selection, we split the activation dataset into training ($70\%$) and validation ($30\%$) sets using stratified sampling with a fixed random state ($1331$). Probes were implemented as Logistic Regression models via \texttt{scikit-learn}, utilizing a \texttt{StandardScaler} and a \texttt{liblinear} solver with L1 regularization (C=1.0) initialized with the same fixed seed. Importantly, L1 regularization in this layer-probing stage is employed solely to mitigate overfitting in the high-dimensional activation space and improve probe generalization. The actual isolation of sparse neurons is handled exclusively by the non-linear \nama{XGBoost} attribution.

\paragraph{Neuron Selection.} For each layer, neuron-level attribution uses an \nama{XGBoost} model \citep{chen:2016} with a \texttt{binary:logistic} objective, trained on all lexicon entries of $\mathcal{D}^{(i)}$ for $100$ boosting rounds with learning rate $\eta = 0.1$, maximum tree depth $3$, and random seed $1331$; all other hyperparameters are left at their library defaults (including $\lambda = 1$ and $\gamma = 0$). Neuron importance is computed with \nama{XGBoost}'s \texttt{gain} importance, i.e., the average split gain over all splits that use the neuron. We retain only the neurons with non-zero gain, which yields a sparse set of neurons for steering.

\subsection{Generation and Hyperparameters}
For generation, we set a constraint of \textbf{at most $128$ new tokens}. To ensure highly stable and reproducible sampling trajectories despite the stochastic nature of the decoding process, all generation runs were executed with a fixed global random seed ($1331$) initialized across \texttt{Python}, \texttt{NumPy}, and \texttt{PyTorch}. The specific sampling parameters were:
\begin{itemize}
    \item \textbf{Llama:} Temperature $0.6, p=0.9$.
    \item \textbf{Qwen:} Temperature $0.6, k=20, p=0.95$.
    \item \textbf{Gemma:} Default settings ($k=64, p=0.95$).
\end{itemize}

The steering intensity $\alpha$ was tuned via grid search on a $50$-prompt tuning split. Based on the highest \textbf{Success Rate} (our primary metric), we selected $\alpha = 0.5$ for monolingual steering across all models. For multilingual steering, we used $\alpha = 0.75$ for \nama{Gemma} and \nama{Qwen}, and $\alpha = 1.0$ for \nama{Llama}.

\subsection{Validation of LLM-as-a-Judge for Multilingual Sentiment}
\label{app:llm_judge_validation}

Evaluating sentiment shifts in open-ended generative text across typologically diverse languages requires a metric capable of capturing complex, context-dependent semantics. While traditional lexicon-based scoring approaches offer a highly interpretable alternative \citep{taboada:2011, liu:2012}, they present severe structural limitations for our specific experimental setup. Our evaluation suite spans 11 languages, including highly agglutinative systems (e.g., Basque, Japanese, Indonesian, Javanese, Sundanese) and morphologically fusional languages (e.g., Estonian, Icelandic). In these typologies, sentiment-bearing roots undergo extensive affixation and internal stem mutations, causing static dictionary engines to suffer from catastrophic surface-form miss rates without flawless, language-specific lemmatizers. 

Furthermore, lexicon approaches are inherently compositionally blind; they aggregate static token weights and struggle to resolve contextual polysemy or implicit sentiment distributed across narrative structures \citep{khoo:2018}. To overcome these limitations, we utilize an LLM-based evaluator (\nama{GPT-4.1}), which natively resolves morphological variants via sub-word tokenization and captures deep contextual dependencies.

To empirically assess the reliability of our evaluation metric and address broader cross-lingual limitations noted in recent literature \citep{xu:2025}, we manually audited the \nama{GPT-4.1} judge across both high-resource (English) and low-resource (Javanese) settings. In both cases, 150 generated instances were sampled (50 per predicted sentiment class), fully randomized, and evaluated with the human annotator blinded to the model's predictions. The English baseline was audited by one of the authors, whereas the Javanese evaluation was conducted by a volunteer native speaker recruited from the authors' network to ensure a rigorous assessment of low-resource linguistic nuances. To ensure a fair comparison, the human annotator evaluated the text using criteria identical to the LLM judge (detailed in Appendix \ref{app:sentiment_prompt}). The annotator provided verbal consent for their labels to be used for research validation; no personal or demographic data was collected.

\begin{table}[ht]
\centering
\small
\begin{tabular}{l | ccc | c}
\toprule
\multicolumn{5}{c}{\textbf{English (High-Resource)}} \\
\midrule
& \multicolumn{3}{c|}{\textbf{Human Annotation (Gold)}} & \\
\textbf{Predicted} & \textbf{Negative} & \textbf{Positive} & \textbf{Neutral} & \textbf{Precision} \\
\midrule
Negative & \textbf{47} & 0 & 3 & 94\% \\
Positive & 0 & \textbf{50} & 0 & 100\% \\
Neutral  & 1 & 1 & \textbf{48} & 96\% \\
\midrule
\multicolumn{4}{r|}{\textbf{Macro-Average}} & \textbf{96.7\%} \\
\midrule
\multicolumn{5}{c}{\textbf{Javanese (Low-Resource)}} \\
\midrule
& \multicolumn{3}{c|}{\textbf{Human Annotation (Gold)}} & \\
\textbf{Predicted} & \textbf{Negative} & \textbf{Positive} & \textbf{Neutral} & \textbf{Precision} \\
\midrule
Negative & \textbf{46} & 1 & 3 & 92\% \\
Positive & 7 & \textbf{36} & 7 & 72\% \\
Neutral  & 9 & 6 & \textbf{35} & 70\% \\
\midrule
\multicolumn{4}{r|}{\textbf{Macro-Average}} & \textbf{78.0\%} \\
\bottomrule
\end{tabular}
\caption{Confusion matrices for the \nama{GPT-4.1} sentiment judge across 150 randomized samples per language.}
\label{tab:confusion_matrices}
\end{table}

The detailed confusion matrices (Table \ref{tab:confusion_matrices}) provide deeper insight into the model's error distribution across resource tiers. While there is an expected performance drop in the low-resource Javanese setting (achieving a macro-average precision of $78$\%), the matrix reveals that errors are predominantly boundary confusions between polar and neutral classes rather than severe polar inversions. For instance, the judge only predicted a positive sentiment for a true negative sample in $7$ out of $150$ instances, maintaining excellent directional precision for negative sentiment shifts ($92$\%). These empirical findings confirm that despite the structural complexities of low-resource languages, the contextual LLM judge provides a highly reliable metric for capturing directional sentiment shifts.


\newpage
\section{Prompts and Dataset Topics}
\label{app:prompts_and_topics}

\subsection{Prompts for Data Generation}
We utilized specific prompts to generate the categories, topics, and neutral sentences. For low-resource languages (Hausa, Javanese, Sundanese), a specialized translation prompt was used to ensure stylistic accuracy.

\begin{center}
\fbox{%
    \parbox{\dimexpr\columnwidth-2\fboxsep-2\fboxrule\relax}{%
        \footnotesize
        \textbf{1. Category and Topic Generation Prompt} \\ 
        I would like to conduct an experiment to examine whether a Large Language Model (LLM) can be steered to generate positive or negative continuations from neutral prompts. Please suggest 10 suitable topic categories, each containing 10 broad subcategories that can serve as a basis for constructing neutral prompts. I plan to create 10 prompts for each subcategory; therefore, the subcategories should remain general and widely applicable rather than overly specific. For reference, I prefer the subcategories to be phrased in a style similar to the examples below:

        \begin{itemize}
            \item \textbf{Preferred style (broad and illustrative):}
            \begin{enumerate}
                \item Morning routine (waking up, breakfast, commuting)
                \item House chores (cleaning, cooking, laundry)
                \item Daily commute (bus, train, traffic)
                \item Leisure activities (reading, walking, hobbies)
                \item Communication (texting, calling, chatting)
                \item Time management (schedules, delays, appointments)
                \item Weather and environment (rain, heat, wind)
                \item Personal interactions (neighbors, family, friends)
                \item Weekend activities (shopping, rest, recreation)
                \item Unexpected daily events (forgotten items, late arrivals)
            \end{enumerate}
            \vspace{0.1cm}
            \item \textbf{Less preferred style (too narrow or fragmented):}
            \begin{enumerate}
                \item Morning routines
                \item Commuting (bus, train, car)
                \item Household chores (e.g., laundry, dishes)
                \item Running errands
                \item Grocery shopping
                \item The weather forecast
                \item Receiving mail or a package
                \item Walking in the neighborhood
                \item Preparing a simple meal
                \item The end of the day
            \end{enumerate}
        \end{itemize}
    }%
}
\end{center}
\begin{center}
\fbox{%
    \parbox{\dimexpr\columnwidth-2\fboxsep-2\fboxrule\relax}{%
        \small
        \textbf{2. Neutral Sentence Generation Prompt} \\ 
        I would like to conduct an experiment to examine whether a Large Language Model (LLM) can be steered to generate positive or negative continuations from neutral prompts.
        
        \vspace{0.2cm}
        To prepare the dataset, I need your help in creating these neutral prompts. I already have a selection of topics from the [category] category, consisting of the following topics:\\
        {[topics]}
        
        \vspace{0.2cm}
        Please generate [amount] neutral prompts for each topic. First, write each prompt in English, then translate it into [languages].
        
        \vspace{0.2cm}
        The output should be formatted in JSONL with a unique prompt\_id, category, topic, and language codes, as shown below. The prompt\_id should start from 1 and increment sequentially.
        
        \vspace{0.2cm}
        {\ttfamily\raggedright
        \{``prompt\_id": 1, ``category": ``morning routine", ``topic": ``waking up", ``en": ``The alarm clock went off at 6 AM.", ``id": ``Jam weker berbunyi pukul 6 pagi.", ``de": ``Der Wecker klingelte um 6 Uhr morgens.", ``zh": ``\begin{CJK*}{UTF8}{gbsn}早上6点，闹钟响了。\end{CJK*}"\} \par}
        
        \vspace{0.2cm}
        \footnotesize Note: The topic details (i.e., explanations in parentheses) do not need to be included in the JSONL output.
    }%
}
\end{center}
\begin{center}
\fbox{%
    \parbox{\dimexpr\columnwidth-2\fboxsep-2\fboxrule\relax}{%
        \small
        \textbf{3. Low-Resource Translation Prompt} \\ 
        I am building a parallel dataset for low-resource languages. Below is a list of JSON objects containing English sentences (``en").

        \vspace{0.2cm}
        Please update each JSON object by adding translations for the following keys:
        \begin{enumerate}\small
            \item ``ha": Hausa (Standard).
            \item ``jv": Javanese (Style: Ngoko Lugu/Casual. Strictly NO Krama/Formal language).
            \item ``su": Sundanese (Style: Basa Loma/Familiar. Use daily conversation style, comparable to Javanese Ngoko).
        \end{enumerate}
        
        \vspace{0.2cm}
        Constraints:
        \begin{itemize}\small
            \item Output ONLY the final valid JSONL format.
            \item Do not add any conversational text before or after the data.
            \item Please enclose the final output in a single code block.
        \end{itemize}
        
        \vspace{0.2cm}
        Input Data:\\
        {[JSONL dataset containing English neutral sentences]}
    }%
}
\end{center}

\newpage
\subsection{List of Categories and Topics}
Below is the complete list of $100$ topics categorized by domain. Topics marked with an asterisk (*) were also selected for the preliminary tuning dataset ($50$ sentences).

\begin{description}
    \item[1. Work \& Professional Life] \hfill \\
    Starting the workday*, Team collaboration*, Managing tasks*, Workplace communication, Taking a break, Meetings, Skill development, Performance feedback, Problem-solving*, Ending the workday*.
    
    \item[2. Home \& Household] \hfill \\
    Morning routine*, Household chores*, Meal preparation*, Home maintenance, Running errands, Family interactions, Managing finances*, Pet care, Receiving mail, Evening routine*.

    \item[3. Social \& Relationships] \hfill \\
    Making plans*, Social gatherings*, Casual conversation*, Community involvement, Meeting new people, Family events*, Online social media, Maintaining contact, Discussing topics, Public interactions*.

    \item[4. Travel \& Transport] \hfill \\
    Daily commute*, Planning a trip*, Air travel, Using public transport, Accommodation, Navigating a new place*, Eating out*, Sightseeing, Unexpected travel events*, Returning home.

    \item[5. Health \& Wellness] \hfill \\
    Physical exercise*, Diet and nutrition*, Sleep patterns*, Medical appointments*, Mental well-being*, Personal hygiene, Feeling unwell, Taking medication, Rest and recovery, Health goals.

    \item[6. Education \& Learning] \hfill \\
    Attending a class*, Studying*, Group projects, Taking an exam*, Doing homework, Interacting with instructors*, Using resources, Learning a new skill*, Class registration, School environment.

    \item[7. Technology \& Digital Life] \hfill \\
    Using a smartphone, Working on a computer*, Email management, Software updates, Troubleshooting*, Online shopping, Digital entertainment*, Data management, Digital security*, Video calls*.

    \item[8. Hobbies \& Leisure] \hfill \\
    Creative activities*, Outdoor pursuits*, Consuming media*, Games*, Attending events, Cooking and baking, Crafts, Following sports, Simple relaxation, Weekend plans*.

    \item[9. Commerce \& Finance] \hfill \\
    Grocery shopping*, Retail shopping, Customer service*, Receiving a service, Public services, Comparing products*, Making a large purchase, Subscription services*, Banking tasks, Following a budget*.

    \item[10. Environment \& Surroundings] \hfill \\
    The weather*, Changing seasons, Environmental noise*, Public spaces*, Recycling and waste*, Weather forecast, Natural events, Indoor environment, Local wildlife, The physical landscape*.
\end{description}

\newpage
\section{Sentiment Evaluation Guidelines}
\label{app:sentiment_prompt}
To ensure a rigorous and fair comparison between the automated metric and human evaluation, both the \nama{GPT-4.1} judge and the human annotator evaluated the generated texts using identical task descriptions and sentiment definitions.

\subsection{LLM Judge Prompt}
We used \nama{GPT-4.1} to evaluate the sentiment of the generated texts. The model was provided with a system instruction to act as a precise classifier and a user prompt containing detailed definitions and constraints. The full prompt structure is provided below:
\begin{center}
\fbox{%
    \parbox{\dimexpr\columnwidth-2\fboxsep-2\fboxrule\relax}{%
        \small
        \textbf{Sentiment Evaluation Prompt Template} \\
        \textbf{System Message:} \\
        {\ttfamily You are a precise sentiment analysis classifier. \par}
        
        \vspace{0.2cm}
        \hrule
        \vspace{0.2cm}
        
        \textbf{User Message:} \\
        {\ttfamily\raggedright
        You are an expert linguist acting as a sentiment judge. \par \vspace{0.1cm}
        The user will provide a text that may be in \{language\} or a mix of languages. \par}
        
        \vspace{0.2cm}
        \textbf{Task:} \\
        {\ttfamily\raggedright
        Analyze the text and classify its dominant sentiment based on the definitions below. \par \vspace{0.1cm}
        If the text contains multiple parts (e.g., instructions, questions, and narrative), focus on the sentiment conveyed by the \textbf{main narrative content} or the \textbf{generated statement}, ignoring purely instructional phrasing (like ``Answer:'', ``Choose A/B'', etc.). \par}
        
        \vspace{0.2cm}
        \textbf{Definitions:} \\
        {\ttfamily\raggedright
        - ``Positive'': The text expresses a positive evaluation, such as indicating that the situation, action, or object is good, helpful, safe, beneficial, desirable, comfortable, satisfying, or otherwise favorable. The evaluation may be explicit (e.g., ``I am happy'') or implicit (e.g., ``the process is efficient''). \par \vspace{0.1cm}
        - ``Negative'': The text expresses a negative evaluation, such as indicating that the situation, action, or object is bad, difficult, harmful, inconvenient, frustrating, disappointing, risky, undesirable, or otherwise unfavorable. The evaluation may be explicit (e.g., ``I am angry'') or implicit (e.g., ``the system is confusing''). \par \vspace{0.1cm}
        - ``Neutral'': The text is purely factual, informative, descriptive, or consists of neutral questions/instructions without emotional valence. \par}
        
        \vspace{0.2cm}
        \textbf{Constraint:} \\
        {\ttfamily The answer must be exactly one word from this list: [Positive, Negative, Neutral] \par}
        
        \vspace{0.2cm}
        \textbf{Text to analyze:} \\
        {\ttfamily \{sentence\} \par}
        
        \vspace{0.2cm}
        \textbf{Answer:}
    }%
}
\end{center}

\subsection{Human Annotator Instructions}
The native Javanese human annotator was provided with the exact same task description and definitions as the LLM judge. The only modifications were the removal of system framing (e.g., ``You are a precise sentiment analysis classifier'') and the substitution of the output constraint with spreadsheet-specific data entry instructions. The provided instructions are detailed below:
\begin{center}
\fbox{%
    \parbox{\dimexpr\columnwidth-2\fboxsep-2\fboxrule\relax}{%
        \small
        \textbf{Human Annotator Instructions} \\
        \hrule
        \vspace{0.2cm}
        
        \textbf{Task:} \\
        {\ttfamily\raggedright
        Analyze the text and classify its dominant sentiment based on the definitions below. \par \vspace{0.1cm}
        If the text contains multiple parts (e.g., instructions, questions, and narrative), focus on the sentiment conveyed by the \textbf{main narrative content} or the \textbf{generated statement}, ignoring purely instructional phrasing (like ``Answer:'', ``Choose A/B'', etc.). \par}
        
        \vspace{0.2cm}
        \textbf{Definitions:} \\
        {\ttfamily\raggedright
        - ``Positive'': The text expresses a positive evaluation, such as indicating that the situation, action, or object is good, helpful, safe, beneficial, desirable, comfortable, satisfying, or otherwise favorable. The evaluation may be explicit (e.g., ``I am happy'') or implicit (e.g., ``the process is efficient''). \par \vspace{0.1cm}
        - ``Negative'': The text expresses a negative evaluation, such as indicating that the situation, action, or object is bad, difficult, harmful, inconvenient, frustrating, disappointing, risky, undesirable, or otherwise unfavorable. The evaluation may be explicit (e.g., ``I am angry'') or implicit (e.g., ``the system is confusing''). \par \vspace{0.1cm}
        - ``Neutral'': The text is purely factual, informative, descriptive, or consists of neutral questions/instructions without emotional valence. \par}
        
        \vspace{0.2cm}
        \textbf{Note:} \\
        {\ttfamily\raggedright
        Select the dropdown on the ``Judgement'' column (rightside of the ``Generated Text'' column) to put your judgment. \par}
        
        \vspace{0.2cm}
        {\ttfamily Thank you. \par}
    }%
}
\end{center}

\newpage
\section{Additional Sentiment Distribution Plots}
\label{app:additional_plots}

In this section, we provide detailed sentiment distribution plots that complement the analysis in Section \ref{sec:steering_results}. First, we present the multilingual steering results for \nama{Qwen-3} and \nama{Llama-3.2}. Second, we provide a comparison against a monolingual (English-only) steering baseline for \nama{Gemma-3}.

\subsection{Multilingual Steering Impact on Sentiment Distribution Across Models}
\begin{figure*}[h]
    \centering
    \includegraphics[width=\textwidth]{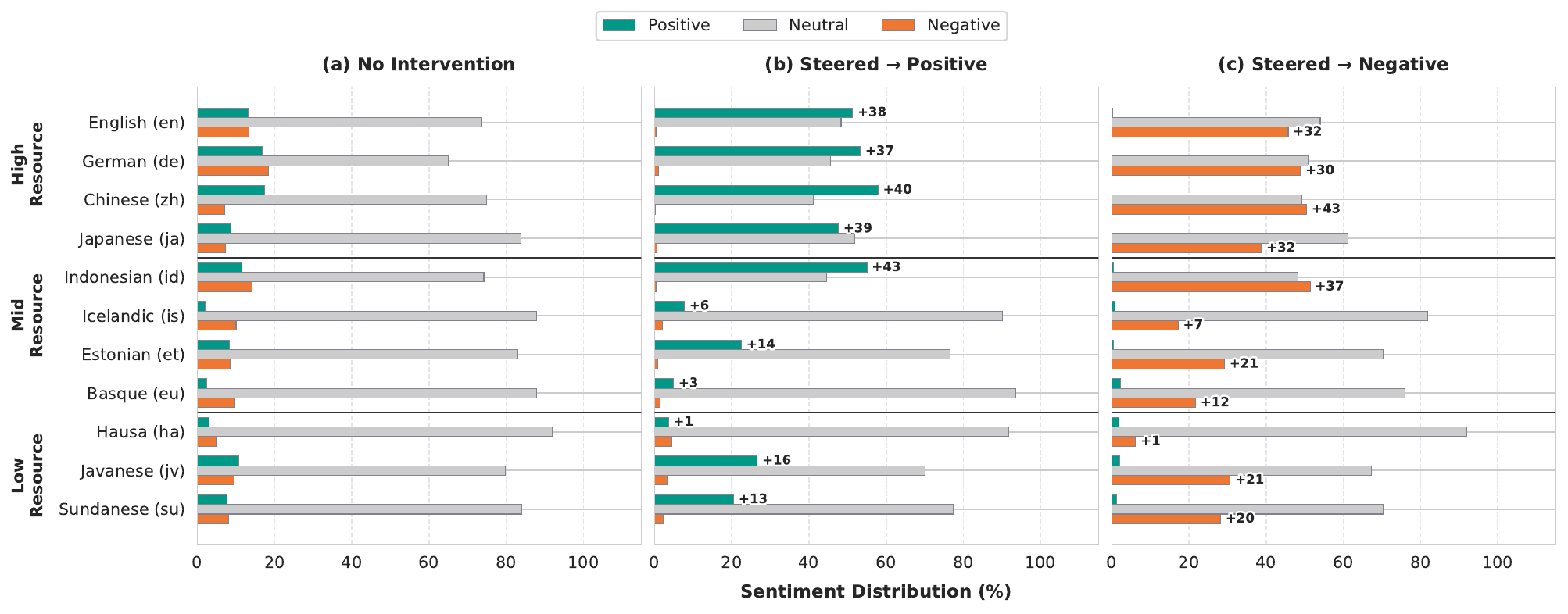}
    \caption{\textbf{Impact of Steering on Sentiment Distribution (Qwen-3).} 
    Steering induces a noticeable shift toward the target sentiment, confirming the method's cross-model applicability. However, Qwen-3 exhibits a stronger tendency to preserve neutral tonality compared to other models, with neutral outputs remaining the dominant class in mid to low resource languages even after intervention.}
    \label{fig:sentiment_dist_qwen}
\end{figure*}
\begin{figure*}[h] 
    \centering
    \includegraphics[width=\textwidth]{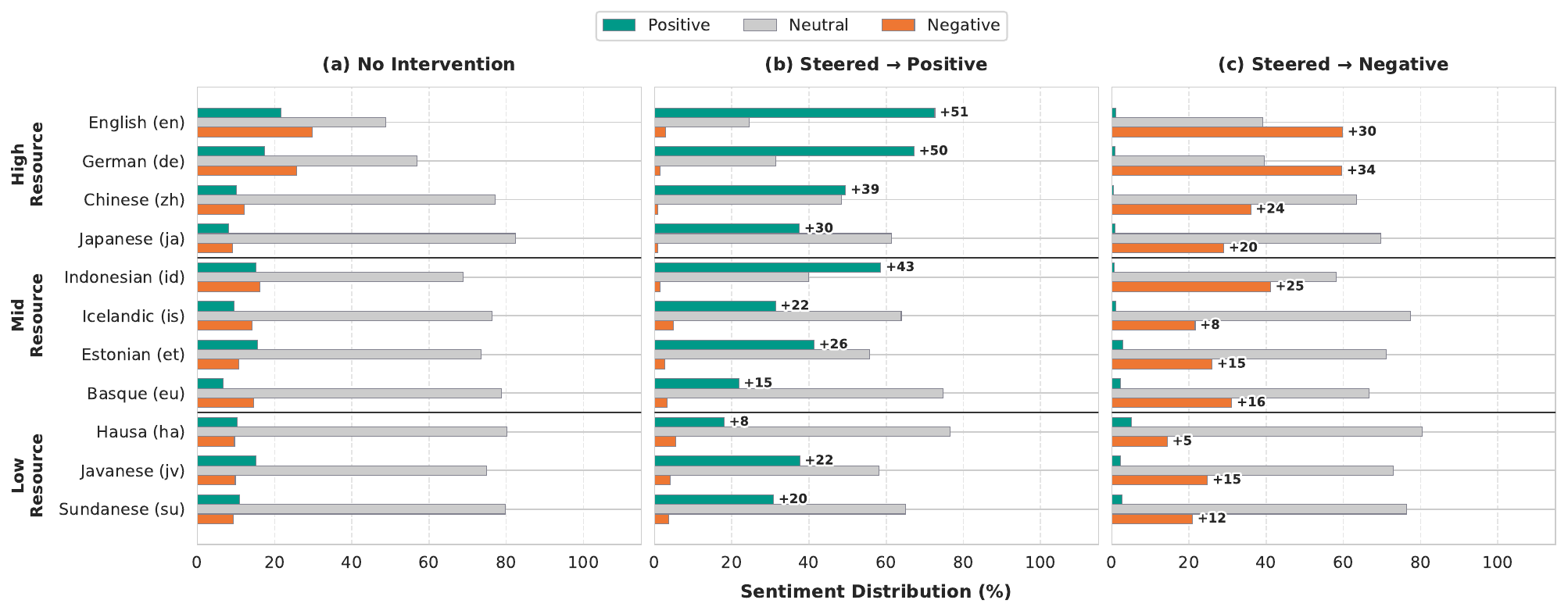} 
    \caption{\textbf{Impact of Steering on Sentiment Distribution (Llama-3.2).} 
    While the steering vectors successfully increase the probability of the target sentiment across all languages, Llama-3.2 retains a higher proportion of neutral outputs compared to Gemma-3, particularly in low-resource languages (e.g., Hausa, Javanese, Sundanese).}
    \label{fig:sentiment_dist_llama}
\end{figure*}

Our results demonstrate that multilingual steering directions consistently influence the sentiment distribution across all three models, shifting the probability mass toward the target sentiment compared to the baseline. However, the magnitude of this effect varies significantly. \nama{Gemma-3} exhibits the highest influence, where the intervention almost completely overrides the default neutral behavior (see Figure~\ref{fig:gemma_sentiment_dist}). In contrast, \nama{Llama-3.2} and \nama{Qwen-3} show a more moderate response, often retaining a significant portion of neutral outputs even after steering.

\subsection{Monolingual (English) Steering Impact on Sentiment Distribution}
\label{app:mono_steering_distribution}
\begin{figure*}[h] 
    \centering
    \includegraphics[width=\textwidth]{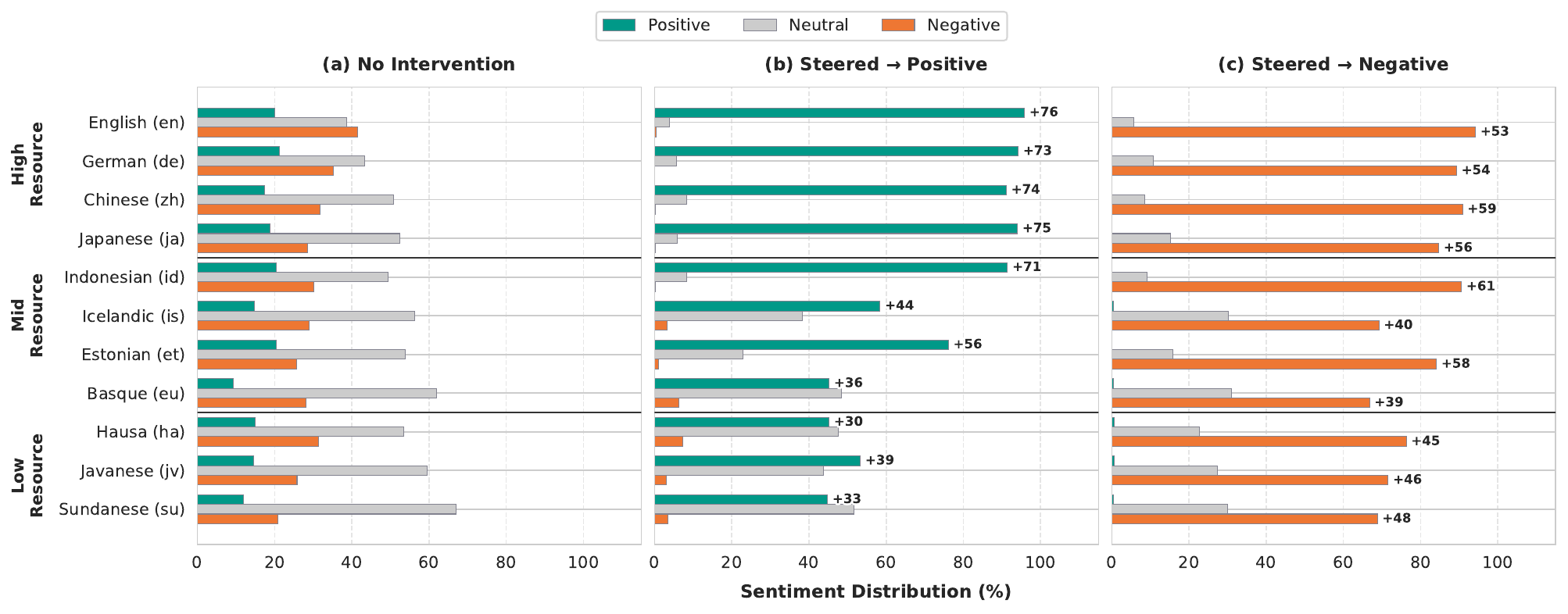}
    \caption{\textbf{Impact of Steering with Monolingual Steering Direction on Sentiment Distribution (Gemma-3).} The distribution of generated sentiments across 11 languages under three conditions: (a) No Intervention (as baseline), (b) Positive Steering, and (c) Negative Steering.}
    \label{fig:gemma_mono_dist}
\end{figure*}

To better understand the distinct contribution of language-specific lexicons, Figure \ref{fig:gemma_mono_dist} visualizes the sentiment distribution shifts for \nama{Gemma-3} using steering directions derived solely from English lexicons. The monolingual directions prove highly effective for high-resource and some mid-resource languages (e.g., German, Indonesian), occasionally matching or even slightly exceeding the multilingual approach, suggesting a strong inherent cross-lingual alignment within \nama{Gemma-3}. However, in specific mid- to low-resource settings (such as Icelandic, Basque, and Sundanese), the English vectors exhibit steeper performance drops and struggle more to overcome the baseline neutral anchor. While the multilingual intervention (Figure \ref{fig:gemma_sentiment_dist}) does not completely eliminate this neutral tendency, it demonstrably mitigates the deterioration, providing a more robust and resilient steering signal for the most resource-constrained languages.

\newpage
\section{Complete Steering Results with SA and LIDR}

\begin{table*}[t]
\centering
\small
\setlength{\tabcolsep}{3.5pt}
\resizebox{\textwidth}{!}{
\begin{tabular}{l c | cccc | cccc | cccc}
\toprule
 & & \multicolumn{4}{c|}{\textbf{Gemma-3}} & \multicolumn{4}{c|}{\textbf{Qwen-3}} & \multicolumn{4}{c}{\textbf{Llama-3}} \\
\textbf{Lang} & \textbf{Res.} & \textbf{SR} & \textbf{SA} & \textbf{LIDR} & \textbf{Config} & \textbf{SR} & \textbf{SA} & \textbf{LIDR} & \textbf{Config} & \textbf{SR} & \textbf{SA} & \textbf{LIDR} & \textbf{Config} \\
\midrule
English (en) & High & $ 95.0 $ & $ 95.0 $ & $ 99.8 $ & all & $ 44.5 $ & $ 54.6 $ & $ 84.1 $ & all & $ 66.4 $ & $ 67.0 $ & $ 98.6 $ & all \\
German (de) & High & $ 90.7 $ & $ 91.8 $ & $ 98.4 $ & all & $ 45.4 $ & $ 52.8 $ & $ 72.1 $ & top-50\% & $ 53.8 $ & $ 64.2 $ & $ 83.0 $ & all \\
Chinese (zh) & High & $ 86.8 $ & $ 91.0 $ & $ 93.5 $ & all & $ 53.4 $ & $ 62.0 $ & $ 82.5 $ & all & $ 34.3 $ & $ 44.5 $ & $ 67.5 $ & all \\
Japanese (ja) & High & $ 80.2 $ & $ 89.3 $ & $ 88.6 $ & all & $ 23.9 $ & $ 31.3 $ & $ 57.5 $ & top-50\% & $ 20.8 $ & $ 38.1 $ & $ 58.2 $ & all \\
\midrule
Indonesian (id) & Mid & $ 82.3 $ & $ 91.0 $ & $ 89.8 $ & all & $ 42.9 $ & $ 47.9 $ & $ 85.1 $ & top-50\% & $ 38.6 $ & $ 51.3 $ & $ 69.1 $ & all \\
Icelandic (is) & Mid & $ 62.3 $ & $ 63.8 $ & $ 97.6 $ & top-50\% & $ 12.1 $ & $ 13.2 $ & $ 95.5 $ & top-50\% & $ 20.0 $ & $ 24.1 $ & $ 89.6 $ & all \\
Estonian (et) & Mid & $ 78.0 $ & $ 80.1 $ & $ 97.5 $ & top-50\% & $ 19.2 $ & $ 22.3 $ & $ 87.1 $ & top-50\% & $ 22.1 $ & $ 29.5 $ & $ 80.2 $ & top-50\% \\
Basque (eu) & Mid & $ 54.7 $ & $ 56.0 $ & $ 96.0 $ & top-50\% & $ 9.2 $ & $ 10.2 $ & $ 92.0 $ & top-50\% & $ 19.6 $ & $ 26.6 $ & $ 79.0 $ & all \\
\midrule
Hausa (ha) & Low & $ 54.6 $ & $ 60.8 $ & $ 90.0 $ & top-50\% & $ 1.8 $ & $ 5.1 $ & $ 61.8 $ & top-5 & $ 9.3 $ & $ 19.8 $ & $ 74.0 $ & top-50\% \\
Javanese (jv) & Low & $ 37.8 $ & $ 62.5 $ & $ 66.0 $ & top-50\% & $ 10.2 $ & $ 30.9 $ & $ 49.1 $ & top-50\% & $ 4.7 $ & $ 34.0 $ & $ 29.9 $ & all \\
Sundanese (su) & Low & $ 25.6 $ & $ 56.8 $ & $ 50.3 $ & top-50\% & $ 4.7 $ & $ 24.8 $ & $ 44.8 $ & top-50\% & $ 7.5 $ & $ 29.1 $ & $ 46.6 $ & all \\
\bottomrule
\end{tabular}
}
\caption{\textbf{Best Results (Mono).} SR: Success Rate, SA: Sentiment Acc, LIDR: Language ID Rate. All metrics in \%.}
\label{tab:results_mono}
\end{table*}

\begin{table*}[t]
\centering
\small
\setlength{\tabcolsep}{3.5pt}
\resizebox{\textwidth}{!}{
\begin{tabular}{l c | cccc | cccc | cccc}
\toprule
 & & \multicolumn{4}{c|}{\textbf{Gemma-3}} & \multicolumn{4}{c|}{\textbf{Qwen-3}} & \multicolumn{4}{c}{\textbf{Llama-3}} \\
\textbf{Lang} & \textbf{Res.} & \textbf{SR} & \textbf{SA} & \textbf{LIDR} & \textbf{Config} & \textbf{SR} & \textbf{SA} & \textbf{LIDR} & \textbf{Config} & \textbf{SR} & \textbf{SA} & \textbf{LIDR} & \textbf{Config} \\
\midrule
English (en) & High & $ 94.3 $ & $ 94.4 $ & $ 99.7 $ & all & $ 45.1 $ & $ 48.5 $ & $ 91.6 $ & all & $ 65.8 $ & $ 66.2 $ & $ 98.9 $ & all \\
German (de) & High & $ 90.5 $ & $ 91.4 $ & $ 98.5 $ & all & $ 46.1 $ & $ 51.1 $ & $ 78.8 $ & top-50\% & $ 53.4 $ & $ 63.4 $ & $ 82.0 $ & all \\
Chinese (zh) & High & $ 86.2 $ & $ 89.0 $ & $ 95.0 $ & all & $ 48.0 $ & $ 54.2 $ & $ 85.8 $ & all & $ 33.1 $ & $ 42.8 $ & $ 71.5 $ & all \\
Japanese (ja) & High & $ 76.8 $ & $ 85.9 $ & $ 87.8 $ & all & $ 23.5 $ & $ 43.2 $ & $ 43.3 $ & all & $ 18.9 $ & $ 33.2 $ & $ 60.1 $ & all \\
\midrule
Indonesian (id) & Mid & $ 82.0 $ & $ 91.0 $ & $ 88.9 $ & all & $ 46.8 $ & $ 53.2 $ & $ 83.2 $ & all & $ 37.0 $ & $ 49.9 $ & $ 69.8 $ & all \\
Icelandic (is) & Mid & $ 66.8 $ & $ 68.8 $ & $ 96.6 $ & all & $ 10.9 $ & $ 12.4 $ & $ 94.3 $ & all & $ 23.1 $ & $ 26.5 $ & $ 91.6 $ & top-50\% \\
Estonian (et) & Mid & $ 77.8 $ & $ 80.2 $ & $ 96.8 $ & all & $ 23.5 $ & $ 25.9 $ & $ 90.8 $ & top-50\% & $ 25.4 $ & $ 33.7 $ & $ 78.8 $ & top-50\% \\
Basque (eu) & Mid & $ 60.7 $ & $ 61.9 $ & $ 95.2 $ & all & $ 12.2 $ & $ 13.3 $ & $ 91.3 $ & all & $ 20.4 $ & $ 26.5 $ & $ 84.7 $ & top-50\% \\
\midrule
Hausa (ha) & Low & $ 58.9 $ & $ 64.4 $ & $ 90.1 $ & top-50\% & $ 2.1 $ & $ 4.9 $ & $ 60.9 $ & top-5 & $ 9.6 $ & $ 16.2 $ & $ 75.7 $ & top-5 \\
Javanese (jv) & Low & $ 37.9 $ & $ 62.7 $ & $ 64.5 $ & top-50\% & $ 9.8 $ & $ 28.6 $ & $ 52.3 $ & top-50\% & $ 4.2 $ & $ 31.2 $ & $ 33.5 $ & top-50\% \\
Sundanese (su) & Low & $ 27.2 $ & $ 63.7 $ & $ 46.1 $ & all & $ 5.7 $ & $ 24.3 $ & $ 48.0 $ & top-50\% & $ 5.9 $ & $ 25.9 $ & $ 47.9 $ & top-50\% \\
\bottomrule
\end{tabular}
}
\caption{\textbf{Best Results (Multi).} SR: Success Rate, SA: Sentiment Acc, LIDR: Language ID Rate. All metrics in \%.}
\label{tab:results_multi}
\end{table*}
Tables \ref{tab:results_mono} and \ref{tab:results_multi} illustrate the comparative strengths of \nama{Gemma-3}, \nama{Qwen-3}, and \nama{Llama-3.2}. Notably, \nama{Gemma-3} maintains significantly higher Success Rates (SR) and Sentiment Accuracy (SA) in low-resource categories (e.g., Hausa, Javanese) compared to its counterparts in both monolingual and multilingual environments.

\newpage
\section{Statistical Significance and Decoding Variance}
\label{app:stats_significance}
In this section, we provide a rigorous statistical analysis of our evaluation results to confirm that the observed performance deltas between multilingual and monolingual steering directions are statistically significant and not artifacts of stochastic decoding variance. We also provide a granular analysis of model behavior in low-resource regimes.

\subsection{Stochastic Decoding and Corpus-Level Stability}
To evaluate the impact of our steering interventions, our experimental pipeline utilizes a massive sample footprint. For each language condition, we conduct $N = 2,000$ independent generation trials (derived from $1,000$ distinct prompts $\times$ 2 target sentiments). 

While decoding parameters remain stochastic, evaluating across thousands of distinct, independent generation episodes leverages the Law of Large Numbers. Local, token-level variance effectively averages out at the corpus level. This stabilizes the observed success rates (SR), allowing them to closely converge toward the true expected performance of the steered model and minimizing the impact of seed-to-seed variance. 

\subsection{Macro- and Micro-Level Statistical Significance}
To verify both the global and localized validity of our approach, we conducted Two-Sample Proportion $Z$-Tests to ensure that the observed performance deltas between multilingual and monolingual steering directions are mathematically meaningful, rather than products of stochastic decoding variance.

\paragraph{Macro-Level Aggregation.} 
First, we aggregated the steering success rates by language resource level across all target models. Because each language tier aggregates multiple balanced conditions, the sample sizes provide immense statistical power ($n=8{,}000$ trials for High/Mid tiers; $n=6{,}000$ trials for the Low tier). Table~\ref{tab:significance_macro} contrasts the Multilingual steering direction against the Monolingual (English) baseline, reporting the success rate for each condition, their difference ($\Delta$~SR, in percentage points), and the corresponding two-tailed test statistics.

\begin{table}[t]
\centering
\footnotesize
\setlength{\tabcolsep}{3pt}
\begin{tabular}{@{}lrrrrr@{}}
\toprule
\textbf{Model} & \textbf{Multi SR} & \textbf{Mono SR} & \textbf{$\Delta$ SR} & \textbf{$Z$-score} & \textbf{$p$-value} \\
\midrule
\multicolumn{6}{@{}l}{\textit{High-resource} ($4$ langs, $n=8{,}000$)} \\
Gemma-3   & $86.95$ & $88.17$ & $-1.22$ & $-2.34$ & $\mathbf{0.0194}$ \\
Qwen-3    & $40.67$ & $41.80$ & $-1.13$ & $-1.45$ & $0.1466$ \\
Llama-3.2 & $42.80$ & $43.83$ & $-1.03$ & $-1.31$ & $0.1886$ \\
\midrule
\multicolumn{6}{@{}l}{\textit{Mid-resource} ($4$ langs, $n=8{,}000$)} \\
Gemma-3   & $71.83$ & $69.33$ & $+2.50$ & $+3.47$ & $\mathbf{0.0005}$ \\
Qwen-3    & $23.35$ & $20.85$ & $+2.50$ & $+3.81$ & $\mathbf{0.0001}$ \\
Llama-3.2 & $26.48$ & $25.08$ & $+1.40$ & $+2.02$ & $\mathbf{0.0429}$ \\
\midrule
\multicolumn{6}{@{}l}{\textit{Low-resource} ($3$ langs, $n=6{,}000$)} \\
Gemma-3   & $41.33$ & $39.33$ & $+2.00$ & $+2.23$ & $\mathbf{0.0255}$ \\
Qwen-3    & $\phantom{0}5.87$ & $\phantom{0}5.57$ & $+0.30$ & $+0.71$ & $0.4791$ \\
Llama-3.2 & $\phantom{0}6.57$ & $\phantom{0}7.17$ & $-0.60$ & $-1.30$ & $0.1938$ \\
\bottomrule
\end{tabular}
\caption{Two-sample proportion $Z$-tests comparing multilingual vs.\ monolingual (English) steering success rates (SR, \%) by language resource tier. Bold $p$-values indicate rejection of the null hypothesis at $\alpha = 0.05$. Positive $\Delta$~SR favors the multilingual direction; negative favors the monolingual baseline.}
\label{tab:significance_macro}
\end{table}

The aggregate data demonstrate a distinct performance pattern: as language resource tiers drop from High to Mid/Low, the multilingual lexicon yields regular and mathematically meaningful advantages. We reject the null hypothesis across the board for the Mid-resource tier, and for \nama{Gemma-3} in the Low-resource tier, where the base model possesses sufficient capacity to respond to cross-lingual steering.
Notably, the sole significant result favoring the monolingual baseline occurs for \nama{Gemma-3} in the High-resource tier ($\Delta$~SR $= -1.22$ pp, $p = 0.0194$). We interpret this reversal as consistent with our broader findings on architectural dynamics: where monolingual representations are already highly optimized, the multilingual lexicon direction introduces slight multilingual ``noise'' rather than a reinforcement signal, resulting in a marginal performance penalty.

\paragraph{Micro-Level Verification (Case Study: \nama{Gemma-3} Hausa).}
To confirm that the gains observed in viable low-resource models are mathematically sound at the individual language level, we performed a localized two-proportion $z$-test specifically on the \nama{Gemma-3} Hausa results ($N = 2,000$ trials per condition):

\begin{itemize}
    \item \textbf{Multilingual Baseline Success:} $1,178 / 2,000$ ($58.9\%$)
    \item \textbf{Monolingual Baseline Success:} $1,092 / 2,000$ ($54.6\%$)
\end{itemize}

This +4.3 pp margin yields a critical value of $z=2.7447$, corresponding to a two-tailed p-value of $0.0060$. This establishes that when baseline capacity thresholds are met, the multilingual lexicon's advantage is substantial and robust, clearing the $p < 0.05$ threshold by a wide margin, proving that the observed low-resource gains are not merely artifacts of stochastic decoding.

\subsection{The Low-Resource Performance Spectrum in Latent Space}
While the statistical significance holds firmly where models exhibit baseline capacity, individual language conditions in extreme low-resource regimes occasionally exhibit marginal absolute performance differences (e.g., $+0.1$ to $+0.3$ percentage points). Rather than indicating methodological failure, our analysis shows these variations reflect a clear architectural spectrum of base model capacity:

\begin{itemize}
    \item \textbf{The Floor Effect (e.g., \nama{Qwen-3} Hausa):} At a success rate of just $1.8\%$, \nama{Qwen-3} experiences a foundational downstream alignment failure in Hausa. Crucially, its baseline Language Identification Rate remains high (LIDR $> 60\%$), indicating the model recognizes the linguistic context. However, the underlying semantic features are not sufficiently robust within the latent space to support targeted linear interventions. In this extreme regime, the steering vector lacks a mature geometric representation to act upon, resulting in statistically insignificant variance (e.g., a marginal $+0.3$ pp delta).
    \item \textbf{Neutral Transfer Equivalence (e.g., \nama{Gemma-3} Javanese):} In contrast to the floor effect, \nama{Gemma-3} possesses functional baseline capacity in Javanese, maintaining a stable success rate of roughly $37.8\%$. The marginal $+0.1$ pp delta represents a regime of \textit{representation equivalence}. In this condition, both the cross-lingual multilingual vector and the monolingual vector achieve matching directional utility in the residual stream, resulting in stable, net-neutral transfer.
    \item \textbf{Positive Cross-Lingual Transfer (e.g., \nama{Gemma-3} Hausa):} When a base model crosses the necessary capacity threshold to respond robustly to interventions in a low-resource language, the architectural advantage of the multilingual lexicon becomes highly visible, delivering the statistically significant boost of $+4.3$ percentage points verified in the previous section.
\end{itemize}

\newpage
\section{Latent Space Geometry and Steering Norms}
\label{app:steering_norms}
\begin{figure*}[h]
    \centering
    \includegraphics[width=\textwidth]{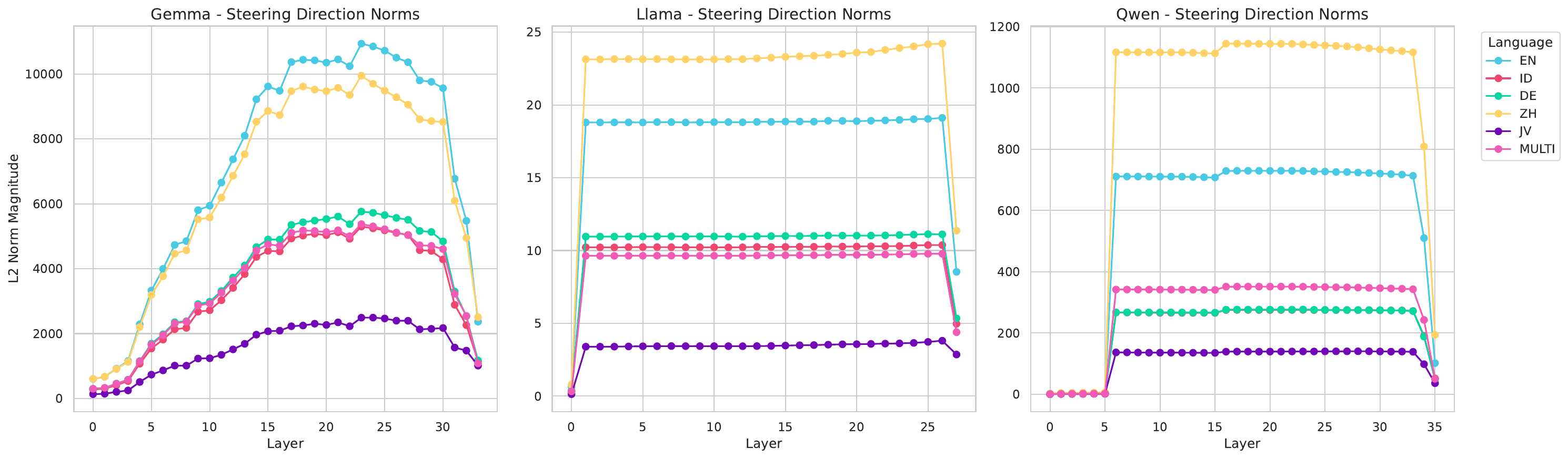}
    \caption{Comparison of $L_2$ norm magnitudes for sentiment steering directions (all neurons) across \nama{Gemma-3}, \nama{Llama-3.2}, and \nama{Qwen-3}. Monolingual directions for German, Chinese, Indonesian, and Javanese are extracted and plotted solely to analyze the latent geometry and are not used for injection.}
    \label{fig:norms_comparison}
    
    \vspace{8pt} 
\end{figure*}
\begin{figure*}[t]
    \centering
    \includegraphics[width=\textwidth]{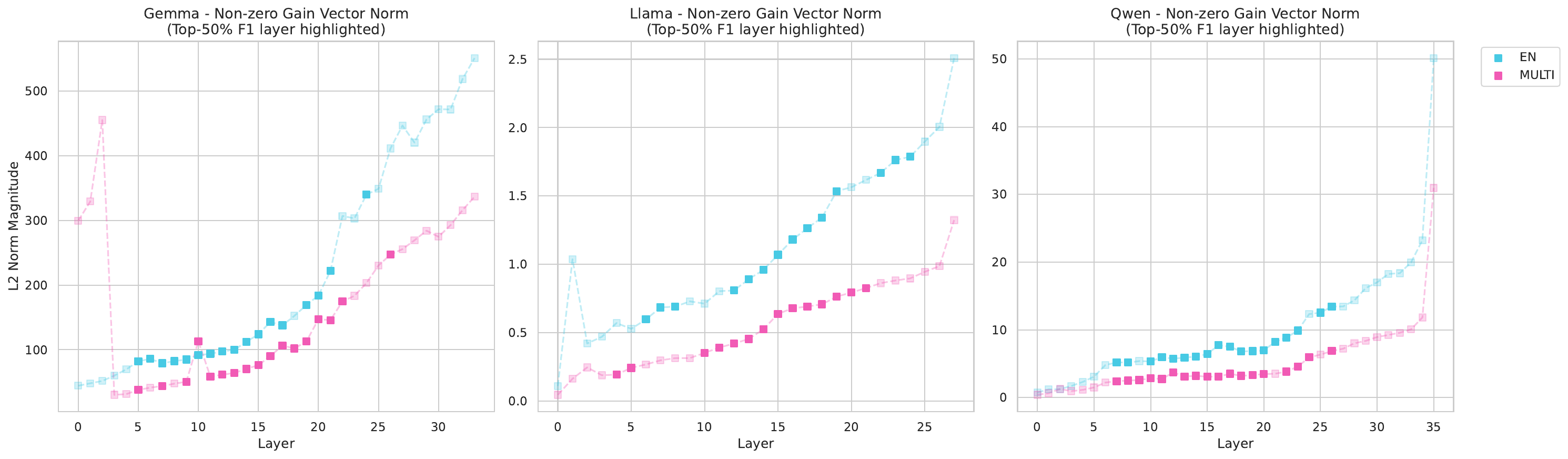}
    \caption{
        $L_2$ norm magnitude of the nonzero-gain steering vectors across layers for \nama{Gemma-3}, \nama{Llama-3.2}, and \nama{Qwen-3}, comparing English and multilingual steering directions. Only neurons with non-zero \nama{XGBoost} gain are included in the norm computation, reflecting the actual magnitude of intervention applied at inference time.
        Opaque markers denote layers in the top 50\% by F1 score of the corresponding linear probe; faded markers denote layers outside this top half.
    }
    \label{fig:nonzero_gain_norm_f1}
\end{figure*}
To understand the physical ``intensity'' of steering interventions, we calculate the $L_2$ norm of the steering direction vectors at each layer. As shown in Figure~\ref{fig:norms_comparison} and \ref{fig:nonzero_gain_norm_f1}, the models exhibit vastly different latent space scales and layer-wise dynamics.
\subsection{Scale and Specialization Ceiling}
The absolute magnitude of steering vectors varies by three orders of magnitude between architectures, despite their comparable hidden dimensions ($2560$ neurons for \nama{Gemma-3} and \nama{Qwen-3}; $3072$ for \nama{Llama-3.2}). Specifically, \nama{Gemma-3} operates in a high-variance regime with norms peaking over $10,000$ in its middle layers, whereas \nama{Qwen-3} exhibits a moderate scale (peaking near $1,150$) and \nama{Llama-3.2} utilizes a highly compressed space with norms remaining below $25$. In all models, the norm magnitude is positively correlated with language resource levels and model specialization. The Chinese (\texttt{zh}) steering direction consistently defines the ``norm ceiling'' in \nama{Qwen-3} ($\sim 1,150$) and \nama{Llama-3.2} ($\sim 24$), followed closely by English (\texttt{en}). This suggests that the model’s most specialized features are encoded with the highest numerical weights. 

The significant disparity in steering effectiveness between models can be partially attributed to this internal scaling of their latent spaces. As illustrated in Figure~\ref{fig:norms_comparison}, the absolute magnitude of steering vectors in \nama{Gemma-3} is several orders of magnitude larger than those in \nama{Llama-3.2} and \nama{Qwen-3}. This suggests a ``stretched'' or high-variance latent space where semantic features are represented with massive numerical weights. Such high-intensity representations may make specific features easier for the model's attention mechanisms to prioritize during token generation, explaining its responsiveness to sentiment steering.

\paragraph{Isolating Effective Intervention Magnitude.}
While Figure \ref{fig:norms_comparison} illustrates the total latent scale, evaluating the non-zero gain vectors (Figure \ref{fig:nonzero_gain_norm_f1}) reveals the concentrated ``active'' magnitude of the steering interventions. Filtering for non-zero gains reduces the $L_2$ norm magnitudes drastically across all architectures: \nama{Gemma-3} drops from $>10,000$ to peak values around $500$. Similarly, \nama{Qwen-3} reduces from $\sim 1,150$ to $<50$, and \nama{Llama-3.2} drops from $\sim 25$ to $<2.5$. Despite this massive reduction in absolute magnitude, the hierarchical relationship between the architectures (\nama{Gemma} $>$ \nama{Qwen} $>$ \nama{Llama}) remains strictly preserved, confirming that \nama{Gemma-3}'s high-variance regime is a fundamental architectural property rather than an artifact of inactive neurons.  

\subsection{The Compactness of Low-Resource Representations}
In stark contrast to the high norm ceilings of English and Chinese, the Javanese (\texttt{jv}) direction possesses the smallest norm magnitude across all architectures: below $5$ in \nama{Llama-3.2} and below $200$ in \nama{Qwen-3}. This suggests that low-resource languages occupy a physically more compact region of the latent space. We hypothesize that the ``linguistic collapse'' observed at high steering intensities for Javanese occurs because external vectors (often scaled to high-resource norms like English) effectively ``over-steer'' these compact clusters, pushing activations far beyond their natural boundaries.
\subsection{Layer-wise Dynamics}
The models demonstrate distinct temporal profiles for sentiment representation:
\begin{itemize}
    \item Static Encoding (\nama{Llama} \& \nama{Qwen}): These models exhibit a ``step-function'' behavior. In \nama{Llama-3.2}, the sentiment direction is fully established by Layer 1, and in \nama{Qwen-3} by Layer $6$, remaining static until the final layers. This implies sentiment is treated as a global, invariant attribute.
    \item Dynamic Progression (\nama{Gemma}): \nama{Gemma-3} shows a ``bell-curve'' progression where the norm builds through the initial layers, peaks between Layers $15$ and $25$, and tapers off toward the output. This indicates a more dynamic semantic refinement process.
\end{itemize}

While dense vectors exhibit static ``step-functions'' (\nama{Llama}, \nama{Qwen}) or ``bell-curves'' (\nama{Gemma}) as shown in Figure \ref{fig:norms_comparison}, isolating the non-zero gain vectors reveals a different temporal dynamic. As illustrated in Figure \ref{fig:nonzero_gain_norm_f1}, the non-zero gain magnitude steadily increases in the deeper layers across all three models. This indicates that while the global sentiment direction may be established early in the network, the concentrated, manipulable weights required to effectively steer that sentiment accumulate and scale up significantly toward the later layers.

\subsection{Optimal Layer Configuration Asymmetry Between Steering Directions.} 
Another consequence of this norm disparity is visible in the optimal layer configurations reported in Tables \ref{tab:results_mono} and \ref{tab:results_multi}. Across nearly all models and languages, the best-performing monolingual (English) steering direction requires fewer layers (e.g., top-$5$ or top-$50$\%) to reach its optimal success rate, whereas multilingual directions more often require intervention across all layers. This phenomenon is visually supported by both our dense and sparse norm analyses. As shown in Figure \ref{fig:norms_comparison}, the English steering direction consistently exhibits a substantially larger $L_2$ norm than the multilingual direction at comparable layers. This disparity is further corroborated by the non-zero gain vectors (Figure \ref{fig:nonzero_gain_norm_f1}), which explicitly map the optimal intervention zones. Within these optimal middle-to-late layers, the concentrated, non-zero magnitude of the English direction (blue) remains distinctly higher than that of the multilingual direction (pink) across all three architectures. Because the magnitude of the injected monolingual perturbation is definitively larger, fewer layers of intervention are needed to displace the model’s activations sufficiently to shift sentiment. Conversely, the multilingual direction, being comparatively dampened in norm, requires broader layer-wise application to achieve its optimal steering efficacy.

Interestingly, this norm-layer relationship does not hold uniformly in \nama{Llama-3.2} for several languages (Icelandic, Basque, Hausa, Javanese, Sundanese), where the optimal monolingual configuration in fact requires more layers than the multilingual counterpart. This anomaly may be attributed to the relatively high cross-lingual vector alignment observed in \nama{Llama-3.2} between English and low-resource language (see Section~\ref{app:vector_alignment}). For instance, the gap between English–Javanese similarity and Multilingual–Javanese similarity is markedly smaller in \nama{Llama-3.2} ($0.970 - 0.968 = 0.002$) than in \nama{Gemma-3} ($0.987 - 0.983 = 0.004$) or \nama{Qwen-3} ($0.937 - 0.912 = 0.025$). Because the English steering direction in \nama{Llama-3.2} is already nearly as aligned with low-resource languages like Javanese as the multilingual consensus direction is, monolingual injection is comparatively less destructive than in the other architectures. Consequently, it can tolerate broader layer-wise application without triggering the linguistic collapse observed elsewhere. 

\section{Cross-Lingual Vector Alignment}
\label{app:vector_alignment}
\begin{figure}[H]
     \centering
     \begin{subfigure}[b]{\linewidth}
         \centering
         \includegraphics[width=0.9\textwidth]{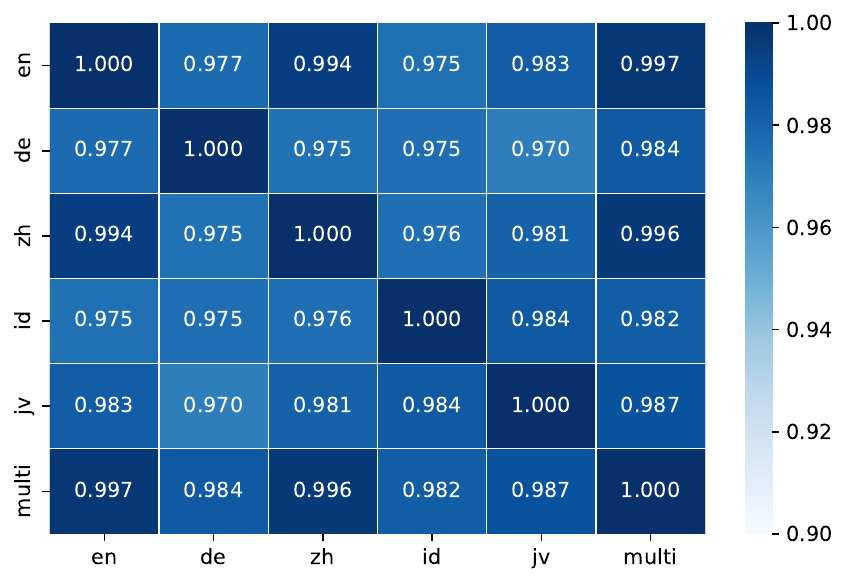}
         \caption{Gemma-3}
         \label{fig:heatmap_gemma}
     \end{subfigure}
     \hfill
     \begin{subfigure}[b]{\linewidth}
         \centering
         \includegraphics[width=0.9\textwidth]{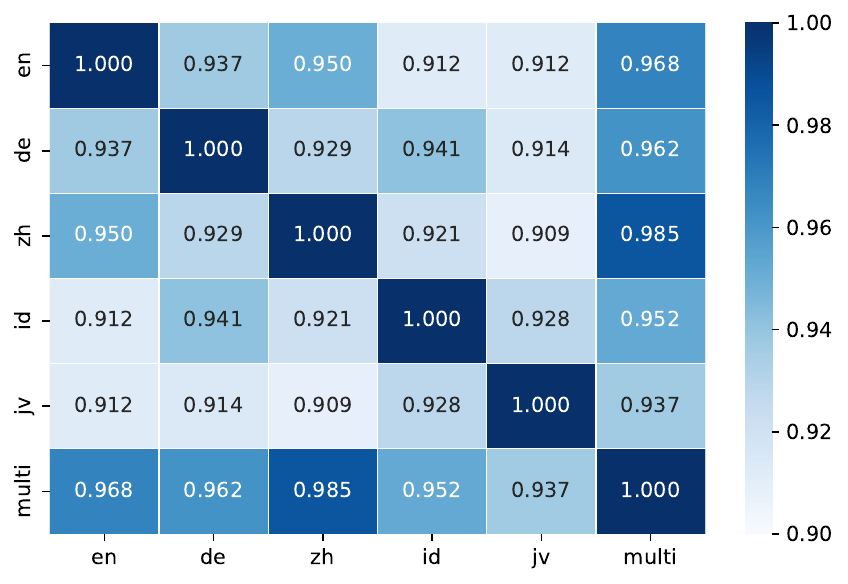}
         \caption{Qwen-3}
         \label{fig:heatmap_qwen}
     \end{subfigure}
     \hfill
     \begin{subfigure}[b]{\linewidth}
         \centering
         \includegraphics[width=0.9\textwidth]{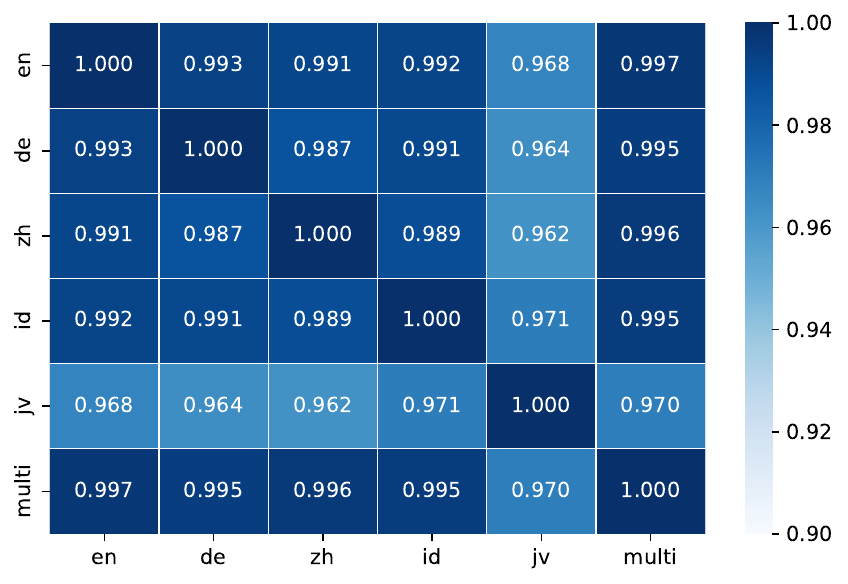}
         \caption{Llama-3.2}
         \label{fig:heatmap_llama}
     \end{subfigure}

    \caption{Cosine similarity heatmaps between monolingual and multilingual steering directions across three architectures. The displayed values represent the average cosine similarity across all layers.}
    \label{fig:combined_heatmaps}

    \vspace{8pt}
\end{figure}
\subsection{The Universal Sentiment Axis}
All three models exhibit high cross-lingual alignment, suggesting the existence of a ``universal'' sentiment axis within their latent spaces.
\begin{itemize}
    \item \nama{Gemma-3} maintains a high ``consensus'' direction, where the multilingual vector's similarity to all anchor languages ranges between $0.982$ and $0.997$.   
    \item \nama{Qwen-3} shows slightly more internal divergence, particularly regarding Chinese (\texttt{zh}), which maintains a $0.985$ similarity to the multilingual vector, while the alignment for Javanese drops to $0.937$.
    \item \nama{Llama-3.2} demonstrates near-perfect uniformity, with the Multilingual (\texttt{multi}) vector achieving a cosine similarity of $>0.99$ with English, German, Chinese, and Indonesian. Even the low-resource Javanese direction remains highly aligned at $0.970$. 
\end{itemize} 

\subsection{Multilingual Dampening Effect}
The similarity data supports our ``consensus'' hypothesis regarding the robustness of multilingual steering. By aggregating diverse lexicons, the resulting vector moves toward a shared centroid, effectively dampening the high-norm ``sharpness'' characteristic of monolingual directions. This effect is most pronounced in \nama{Qwen-3}/Chinese, where the specialized, high-intensity monolingual signal is diluted by the lower-norm profiles of the broader lexicon. While this dampening enhances stability across the ``long tail'' of languages, it accounts for the observed Specialization Penalty in high-resource anchors, where the reduced vector magnitude fails to meet the model's native representational threshold.


\section{Establishing the Causal Relevance of Targeted Intervention Sites}
\label{app:causal_verification}

\subsection{Experimental Setup}
As briefly introduced in Section~\ref{subsec:feature_attribution}, we performed a causal intervention experiment to verify that our isolated layers and neurons actively govern model generation rather than inducing arbitrary changes or gibberish output. We conducted this experiment using the multilingual direction across two low-resource setups: \nama{Gemma-Hausa} and \nama{Qwen-Javanese}. We randomly sampled 50 prompts from our broader 1,000-prompt evaluation set and steered them toward both positive and negative targets, yielding 100 paired evaluation instances per model. To isolate the causal effect of our \nama{XGBoost}-selected neurons, we established our proposed method against two rigorous null-hypothesis baselines: 

\begin{itemize}
    \item \textbf{Proposed Method ($H_1$):} Intervene strictly on the top-50\% probe-ranked layers, using only the specific sparse neurons isolated by \nama{XGBoost}.
    \item \textbf{Fully Random Baseline ($H_0^{(1)}$):} Intervene on a randomly sampled $50$\% of the layers across the architecture, using a completely random subset of neuron indices of identical size ($|\mathcal{N}_i|$).
    \item \textbf{Layer-Controlled Baseline ($H_0^{(2)}$):} Intervene on the exact same top-$50$\% probe-ranked layers, but replace the \nama{XGBoost}-selected coordinates with a random subset of neuron coordinates of identical size ($|\mathcal{N}_i|$).
\end{itemize}

\subsection{Metrics and Statistical Significance}

We evaluated causal efficacy using two metrics: \textbf{Sentiment Accuracy} (whether the generated sentiment matches the intended target) and \textbf{Strict Success Rate (SR)} (whether the text matches the target sentiment \textit{and} retains the correct Language ID, ensuring no language-switching or gibberish occurred). Because the experimental outcomes are paired and binary, we applied McNemar's test to evaluate statistical significance against a standard threshold of $p < 0.05$. 

\subsection{Results and Discussion}
\begin{table*}[htbp]
\centering
\vspace{0.2cm}
\resizebox{\textwidth}{!}{
\begin{tabular}{llcccc}
\toprule
\textbf{Model \& Language} & \textbf{Metric} & \textbf{$H_1$ (Ours)} & \textbf{$H_0^{(1)}$} & \textbf{McNemar $p$-value} & \textbf{Significance ($\alpha = 0.05$)} \\
\midrule
\textbf{Gemma-Hausa} & Sentiment Acc. & \textbf{66\%} & 3\% & $3.09 \times 10^{-17}$ & \textbf{Significant} \\
 & Strict SR & \textbf{56\%} & 0\% & $2.78 \times 10^{-17}$ & \textbf{Significant} \\
\midrule
\textbf{Qwen-Javanese} & Sentiment Acc. & \textbf{27\%} & 7\% & $0.0003$ & \textbf{Significant} \\
 & Strict SR & \textbf{9\%} & 4\% & $0.2266$ & Not Significant \\
\bottomrule
\end{tabular}
}
\caption{Proposed Framework ($H_1$) vs. Fully Random Baseline ($H_0^{(1)}$). This baseline evaluates the effect of applying the steering vector to a random subset of layers ($50$\% of all layers) and randomly sampled neuron indices, preserving the exact intervention size ($\vert{}N_i\vert{}$) of the original targeted configuration.}
\label{tab:feat_attribute_full_random}
\end{table*}

\begin{table*}[htbp]
\centering
\vspace{0.2cm}
\resizebox{\textwidth}{!}{
\begin{tabular}{llcccc}
\toprule
\textbf{Model \& Language} & \textbf{Metric} & \textbf{$H_1$ (Ours)} & \textbf{$H_0^{(2)}$} & \textbf{McNemar $p$-value} & \textbf{Significance ($\alpha = 0.05$)} \\
\midrule
\textbf{Gemma-Hausa} & Sentiment Acc. & \textbf{66\%} & 17\% & $5.91 \times 10^{-12}$ & \textbf{Significant} \\
 & Strict SR & \textbf{56\%} & 16\% & $1.51 \times 10^{-9}$ & \textbf{Significant} \\
\midrule
\textbf{Qwen-Javanese} & Sentiment Acc. & \textbf{27\%} & 12\% & $0.0107$ & \textbf{Significant} \\
 & Strict SR & \textbf{9\%} & 6\% & $0.6072$ & Not Significant \\
\bottomrule
\end{tabular}
}
\caption{Proposed Framework ($H_1$) vs. Layer-Controlled Baseline ($H_0^{(2)}$). This baseline controls for layer location by intervening on the exact same top 50\% probe-ranked layers but replacing our \nama{XGBoost}-selected coordinates with random neuron indices.}
\label{tab:feat_attribute_layer_ctrl}
\end{table*}

Table~\ref{tab:feat_attribute_full_random} and Table~\ref{tab:feat_attribute_layer_ctrl} summarize the outcomes against the fully random and layer-controlled baselines, respectively. For both Gemma-Hausa and Qwen-Javanese, our proposed method ($H_1$) significantly outperforms both the fully random ($H_0^{(1)}$) and layer-controlled ($H_0^{(2)}$) baselines in Sentiment Accuracy ($p < 0.05$). By demonstrating that steering via these specific neurons changes model behavior in a predictable, task-relevant manner compared to random controls, we confirm that these representations are causally relevant, addressing concerns regarding correlative probing \citep{ravichander:2021}.

In Gemma-Hausa, our method yields a highly significant improvement in Strict Success Rate ($p < 10^{-8}$). Randomly perturbing neurons or layers frequently broke the Language ID or induced gibberish, whereas targeting our probe-selected neurons successfully flipped the sentiment while keeping the language generation intact. 

For Qwen-Javanese, the strict success rate difference was not statistically significant. This aligns directly with our findings in Table~\ref{tab:results_multi}, which demonstrates that Qwen has an overall Javanese baseline Success Rate of only $9.8$\%. Because the base model inherently struggles to maintain Javanese text generation, any representation steering is bottlenecked by the model's underlying language generation capabilities. Crucially, however, the Sentiment Accuracy remains statistically significant ($p = 0.0107$ and $p = 0.0003$), demonstrating that the targeted features do encode sentiment even when the base model's language constraints limit full generation success.

\section{Isolating the Multilingual Advantage from Vector Norm and Scaling Effects}

Given the high cosine similarity between the monolingual and multilingual vectors as discussed in Section~\ref{app:vector_alignment}, it is critical to determine whether the observed ``multilingual advantage'' is driven by true cross-lingual semantic directions or if it is an artifact of vector norm differences and alpha scaling ($\alpha$). To isolate the geometric impact of the vector direction from these scaling factors, we present controlled experiments executed on \nama{Gemma-3} for Hausa (ha). Hausa serves as a representative low-resource setting where the base model demonstrates robust performance, ensuring a reliable baseline environment to analyze steering dynamics.

\subsection{Structural Divergence in Steering Pipelines}

\begin{figure*}[ht]
    \centering
    \includegraphics[width=\textwidth]{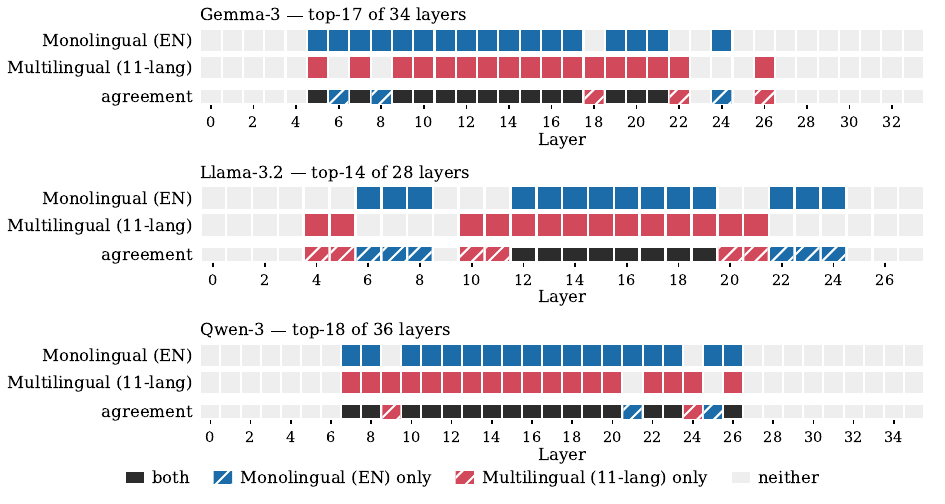}
    \caption{Layer selection divergence across \nama{Gemma-3} (top-17 of 34), \nama{Llama-3.2} (top-14 of 28), and \nama{Qwen-3} (top-18 of 36). Blue and red striped regions indicate layers active exclusively in the Monolingual (EN) and Multilingual (11-lang) setups, respectively, while black indicates agreement.}
    \label{fig:layer_divergence}
\end{figure*}

Our two-stage feature attribution mechanism independently derives the optimal layer configurations and their corresponding active neuron masks for both setups. Consequently, the selected subset of target layers inherently differs. As illustrated in Figure \ref{fig:layer_divergence}, the top-50\% layer sets for the monolingual and multilingual directions exhibit structural divergence across all evaluated models. Specifically, the layer-wise Jaccard similarity is $0.70$ ($14$ of $20$ shared active layers) for \nama{Gemma-3}, $0.40$ ($8$ of $20$ shared) for \nama{Llama-3.2}, and $0.80$ ($16$ of $20$ shared) for \nama{Qwen-3}.

\begin{figure}[h]
    \centering
    \includegraphics[width=\columnwidth]{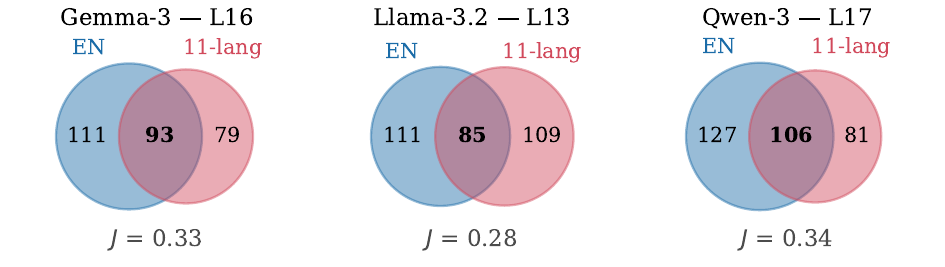}
    \caption{Neuron activation divergence between the Monolingual (EN) and Multilingual (11-lang) directions within the representative middle layer of each model. The minimal overlap across all three architectures confirms that the setups utilize distinct active neuron sub-networks.}
    \label{fig:neuron_divergence}
\end{figure}

Furthermore, even when both setups select the exact same layers, the specific active neurons targeted within those layers diverge substantially. To isolate this neuron-level divergence, we analyze the middle network layer for each model (using $0$-based indexing): Layer $16$ for \nama{Gemma-3}, Layer $13$ for \nama{Llama-3.2}, and Layer $17$ for \nama{Qwen-3} (Figure \ref{fig:neuron_divergence}). We chose these specific middle layers because they are independently identified as active target layers by both the monolingual and multilingual setups. By examining layers where the selection perfectly aligns, we provide an objective baseline to demonstrate that layer-level agreement does not guarantee neuron-level agreement. Within these shared layers, the active neuron masks exhibit minimal overlap (all $J \le 0.34$). This proves that the two setups steer fundamentally distinct internal processing pathways, rather than the multilingual vector acting as a merely scaled version of the English circuit.

\subsection{Norm-Controlled Baselines}
To control for differences in scale, we first rescaled the English vector to match the Euclidean norm of the multilingual vector, 
\begin{equation*}
v_{\text{scaled}} = \frac{\Vert v_{\text{multi}} \Vert_2}{\Vert v_{\text{mono}} \Vert_2} \cdot v_{\text{mono}.}
\end{equation*}

Using this vector, we formalize the validation of the multilingual advantage through the following comparative configurations.
\begin{itemize}
    \item \textbf{Original Multilingual Setup ($H_1$):} Intervene using the multilingual direction vector $v_{\text{multi}}$ within the independently optimized multilingual layer/neuron configuration at $\alpha_{\text{multi}} = 0.75$.
    \item \textbf{Norm/$\alpha$-Controlled Baseline ($H_0^{(A)}$):} Intervene using the rescaled monolingual English vector $v_{\text{scaled}}$ within the native monolingual layer/neuron configuration at $\alpha_{\text{multi}}$.
    \item \textbf{Configuration-Swapped Baseline ($H_0^{(B)}$):} Inject the rescaled monolingual English vector $v_{\text{scaled}}$ directly into the multilingual layer/neuron configuration at $\alpha_{\text{multi}}$.
\end{itemize}

\subsection{Results and Statistical Significance}
We evaluated these baselines against the evaluation dataset consisting of 1,000 unique base prompts. Steering each prompt toward two distinct sentiment targets yields a total of $N = 2000$ paired text evaluations per condition (Table \ref{tab:gain_validation}).

\begin{table}[ht]
\centering
\small
\setlength{\tabcolsep}{4pt}
\begin{tabular}{@{}l c c c c c@{}}
\toprule
\textbf{Setup} & \textbf{Vector} & \textbf{L/N Config} & \textbf{Scale} & \textbf{SA} & \textbf{SR} \\
\midrule
Mono & $v_{\text{mono}}$ & Mono & Tuned & 60.8 & 54.6 \\
$H_0^{(A)}$ & $v_{\text{scaled}}$ & Mono & $\alpha_{\text{multi}}$ & 61.3 & 56.1 \\
$H_0^{(B)}$ & $v_{\text{scaled}}$ & Multi & $\alpha_{\text{multi}}$ & 61.4 & 56.5 \\
\textbf{$H_1$ (Multi)} & \textbf{$v_{\text{multi}}$} & \textbf{Multi} & \textbf{Tuned} & \textbf{64.4} & \textbf{58.9} \\
\bottomrule
\end{tabular}
\caption{Performance of norm-matched and configuration-swapped baselines on \nama{Gemma-3} (Hausa). Metric scores (SA, SR) are in percentages (\%). Statistical significance ($p < 0.05$) across identical prompt instances was achieved for Sentiment Accuracy (SA) against both baselines, but not for Steering Rate (SR).}
\label{tab:gain_validation}
\end{table}
We tested whether the multilingual advantage persists after controlling for vector scale and structural placement. Specifically, we used McNemar's test for paired nominal data, comparing binary outcomes (correct vs.\ incorrect sentiment generation) across identical prompt instances. We set the threshold for statistical significance at $p < 0.05$.

\begin{itemize}
    \item \textbf{$H_1$ vs.\ $H_0^{(A)}$:} This evaluates whether the multilingual direction yields an advantage over a perfectly norm- and scale-matched English vector operating in its native architecture zone. McNemar's test yields $p = 0.0285$. Since $p < 0.05$, we reject $H_0^{(A)}$, confirming that the multilingual advantage is statistically significant over a norm-corrected monolingual setup.
    \item \textbf{$H_1$ vs.\ $H_0^{(B)}$:} This evaluates if the advantage holds when the rescaled monolingual vector is forced into the identical multilingual configuration. McNemar's test yields $p = 0.0374$. Since $p < 0.05$, we reject $H_0^{(B)}$. Even within the cross-lingually optimized pathway, the rescaled English direction underperforms the true multilingual direction.
\end{itemize}

These results demonstrate that while the norm effect is present (rescaling the English vector yields a minor gain in sentiment accuracy ($60.8$\% $\rightarrow$ $61.3$\%)), it cannot fully close the performance gap. In high-dimensional LLM activation spaces, near-collinear cosine similarities can mask orthogonal directional components. Minor directional shifts in this high-dimensional space carry a unique cross-lingual utility that cannot be replicated by scalar adjustments to a monolingual vector. 

\section{Empirical Analysis of Incoherent Outputs}
\label{sec:appendix_linguistic_collapse}

During our investigations, severe linguistic collapse (manifesting as non-words, punctuation loops, or degenerate repetitive sequences) emerged under two distinct experimental contexts. To clarify the failure modes discussed in the main paper, we decouple our preliminary baseline observations from the formal hyperparameter sweep. 

\begin{enumerate}
    \item \textbf{Ablation of Sparse Selection (Dense Intervention):} In Section~\ref{subsec:feature_attribution} (the Sparse Neuron Selection via Non-linear Attribution paragraph), we argue that dense intervention (steering all neurons indiscriminately) injects excessive noise into foundational syntax circuits. As shown in the top half of Table~\ref{tab:linguistic_collapse}, preliminary baselines applying vectors across all neurons consistently resulted in syntax breakdown. Because this dense configuration universally failed to produce evaluable text even at low intensities ($\alpha \le 0.5$), it was excluded from the main evaluation sweep.
    \item \textbf{Magnitude Oversaturation (Figure~\ref{fig:tuning_heatmap} Black Cells):} Even when successfully isolating features via sparse selection (e.g., non-zero gain neurons), excessive steering multipliers force the model outside its operational distribution. In Figure~\ref{fig:tuning_heatmap}, these autoregressive collapses are visually encoded as black cells. We observe that the Monolingual vector shows a higher tendency to trigger collapse at $\alpha \ge 0.75$ across many non-English languages, whereas the Multilingual vector remains comparatively robust.
\end{enumerate}

Table~\ref{tab:linguistic_collapse} compiles representative outputs illustrating both failure regimes. To clearly showcase these degradation patterns, all examples in this table utilize the Monolingual (en) steering direction. Driven by its larger vector norm (see Figure~\ref{fig:norms_comparison}), it is particularly susceptible to both dense-induced collapse and magnitude oversaturation.

\begin{table*}[t]
\centering
\small
\begin{tabular}{>{\centering\arraybackslash}p{2cm} >{\centering\arraybackslash}p{1.6cm} >{\centering\arraybackslash}p{2cm} c c p{3.2cm} p{4cm}}
\toprule
\textbf{Model} & \makecell[c]{\textbf{Layer} \\ \textbf{Config}} & \makecell[c]{\textbf{Neuron} \\ \textbf{Selection}} & \textbf{$\alpha$} & \textbf{Lang} & \multicolumn{1}{c}{\makecell[c]{\textbf{Translated} \\ \textbf{Prompt}}} & \multicolumn{1}{c}{\textbf{Degenerate Output}} \\
\midrule
\multicolumn{7}{c}{\textit{Dense Intervention: Disabling Sparse Selection}} \\
\midrule
Llama3.2-3B & top-5 & all & $0.5$ & en & He turned on his computer. & \texttt{\textbackslash nTheLG-HeCJHG-4HEL-200D-J\allowbreak Question-Question-} \\
\addlinespace
Llama3.2-3B & top-5 & all & $0.3$ & zh & He put on his running shoes. & \texttt{—‑ – – – – – – – – – – – – – – – - - - - - - - - - - - – – - - - - - - - - - - - - - - - - - - - - - - - - - - -.. - - - - - - - - - -........ - ..........................\allowbreak.......} \\
\addlinespace
Qwen3-4B & top-5 & all & $0.2$ & id & The alarm went off at 6:30 AM. & \texttt{,,,,,, ordered of,bbbdon of ordered of of of of, of monn,bonitnakne\allowbreak bonononnebit of, amongonnebnebnebvitit, of are amongnebpitneby\allowbreak bpbnenebb ofbneonbonbbbhnepnenebhon\allowbreak{}itnepnebpitnedithitonnebf\allowbreak{}uccitititbbbb atvnenenelenelepitbononit\allowbreak{}hitneb} \\
\addlinespace
Qwen3-4B & top-5 & all & $0.2$ & de & He looked at his phone screen. & \texttt{\textbackslash n \textbackslash n \textbackslash n \textbackslash n\textbackslash n \textbackslash n\textbackslash n\textbackslash n \textbackslash n \textbackslash n\textbackslash n \textbackslash n \textbackslash n \textbackslash n \textbackslash n\textbackslash n \textbackslash n \textbackslash n\textbackslash n \textbackslash n \textbackslash n\textbackslash n\textbackslash n \textbackslash n \textbackslash n \textbackslash n \textbackslash n \textbackslash n \textbackslash n \textbackslash n \textbackslash n \textbackslash n......... \textbackslash n \textbackslash n \textbackslash n \textbackslash n\textbackslash n \textbackslash n \textbackslash n \textbackslash n \textbackslash n \textbackslash n \textbackslash n \textbackslash n...... \textbackslash nd\textbackslash n\textbackslash nd\textbackslash n... \textbackslash n \textbackslash n \textbackslash n... \textbackslash n \textbackslash n \textbackslash n \textbackslash n \textbackslash n \textbackslash n \textbackslash n.........} \\
\midrule
\multicolumn{7}{c}{\textit{Magnitude Oversaturation: Figure~\ref{fig:tuning_heatmap} Black Cells}} \\
\midrule
Gemma3-4B & all & non-zero gain & 1.0 & jv & She opened the spreadsheet to track expenses. & \texttt{Nalokir sing bisaukak d-r-s-s2 - ing - er or u\textbackslash nAnus\textbackslash n} \\
\addlinespace
Qwen3-4B & all & non-zero gain & 1.0 & jv & The building stood next to the road. & \texttt{Nak Dak Dakakakak, Dak Dak Dak D, D D, D is the, D D, D In, D, D, D, D, D, D, D, D, D, D, D, D, D, D, D, D, D, D, D, D, D, D, D, D, D, D, D, D, D, D, D, D, D, D, D, D; D, D, D; D, D, D; D, D; D, D, D; D, D} \\
\addlinespace
Llama3.2-3B & all & non-zero gain & 1.0 & jv & I sent a text message about meeting up. & \texttt{Itu makan. Bukan. I. But. I. Not. I. I. Not. I. Not. I. Not. Not. I. Not. I. Not. Not. I. Not. Not. I. Not. Not. Not. I. Not. Not. Not. Not. Not. Not. Not. Not. Not. Not. Not. Not. Not. Not. Not. Not. Not. Not. Not. Not. Not. Not. Not. Not. Not. Not. Not. Not. Not. Not. Not. Not. Not. Not. Not. Not. Not.} \\
\bottomrule
\end{tabular}
\caption{Representative degenerate outputs illustrating the two failure modes discussed in the paper. All examples utilize the Monolingual steering vector. Top: Linguistic collapse caused by dense intervention (steering all neurons), excluded from the main sweep. Bottom: Collapse stemming from magnitude oversaturation on sparse features ($\alpha=1.0$), which corresponds directly to the unevaluable black cells in Figure~\ref{fig:tuning_heatmap}.}
\label{tab:linguistic_collapse}
\end{table*}
\end{document}